\documentclass[10pt,twocolumn,letterpaper]{article}

\usepackage{cvpr}              
\usepackage{cuted}    
\usepackage{capt-of}  
\usepackage{pifont}

\definecolor{cvprblue}{rgb}{0.21,0.49,0.74}
\usepackage[pagebackref,breaklinks,colorlinks,allcolors=cvprblue]{hyperref}
\usepackage{booktabs}   
\usepackage{multirow}   
\usepackage{tabularx}   
\usepackage{pifont}     
\usepackage{caption}    
\usepackage{rotating}   
\usepackage{relsize}    
\usepackage{array}        
\usepackage[export]{adjustbox}    
\usepackage{xcolor}       
\usepackage{arydshln} 
\usepackage{graphicx}
\usepackage[margin=1in]{geometry}
\usepackage{array}
\usepackage[table]{xcolor}
\usepackage{makecell}

\usepackage{lmodern}  
\usepackage{hyperref}
\def\paperID{5082} 
\def\confName{CVPR}
\def\confYear{2026}

\newcommand{\ql}[1]{\textcolor{black}{#1}}   

\newcommand{\gx}[1]{\textcolor{black}{#1}}
\newcommand{\yuhang}[1]{\textcolor{black}{#1}}
\title{AirSim360: A Panoramic Simulation Platform within Drone View}

\author{
Xian Ge$^{1}\thanks{, $\ddag$, $\dagger$ indicate equal contribution, project leader and corresponding author.}$,
Yuling Pan$^{1,6*}$,
Yuhang Zhang$^{1*}$,
Xiang Li$^{1}$,
Weijun Zhang$^{1\ddag}$,
Dizhe Zhang$^{1}$,\\
Zhaoliang Wan$^{1}$,
Xin Lin$^{1,3}$,
Xiangkai Zhang$^{1}$,
Juntao Liang$^{1}$,
Jason Li$^{4}$, \\
Wenjie Jiang$^{1}$,
Bo Du$^{2}$,
Ming-Hsuan Yang$^{5}$,
Lu Qi$^{1,2\dagger}$ \\
\\
\small
$^{1}$ Insta360 Research \quad
$^{2}$ Wuhan University \quad
$^{3}$ University of California, San Diego \\
\small
$^{4}$ Nanyang Technological University \quad
$^{5}$ University of California, Merced
$^{6}$ Shenzhen University
}

\begin{document}

\maketitle

\begin{strip}
\centering
\setlength{\fboxsep}{0pt}%
\includegraphics[width=0.99\linewidth]{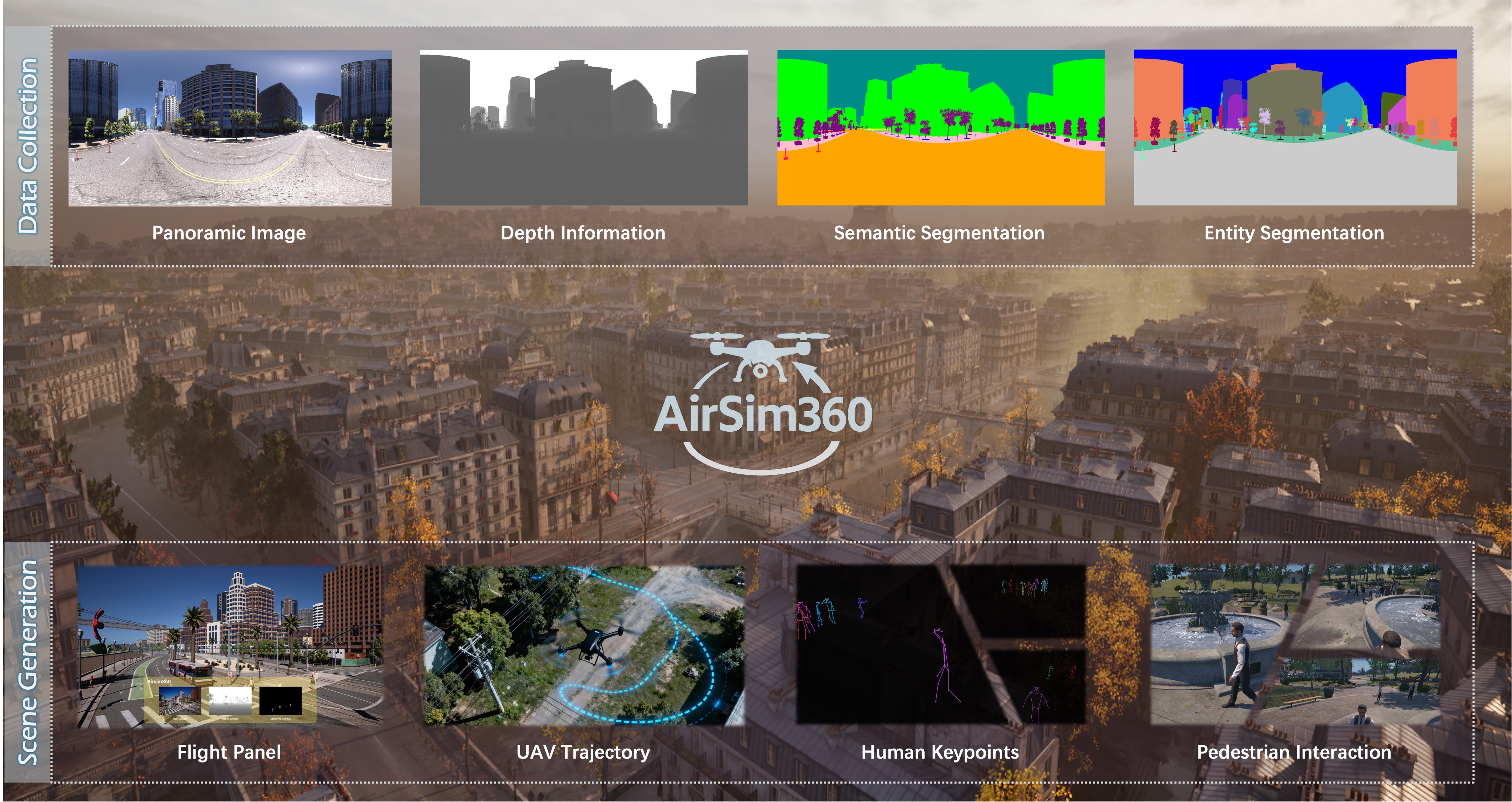}
\captionsetup{hypcap=false}
\captionof{figure}{\yuhang{Overview of Airsim360. This work introduces a panoramic UAV simulation platform based on a cutting-edge rendering engine, enabling closed-loop simulation for omnidirectional aerial systems and offering an integrated toolchain for intelligent data acquisition across diverse flight scenarios.}} 
\vspace{-4mm}         
\end{strip}

\begin{abstract}
\ql{The field of 360-degree omnidirectional understanding has been receiving increasing attention for advancing spatial intelligence. However, the lack of large-scale and diverse data remains a major limitation. In this work, we propose AirSim360, a simulation platform for omnidirectional data from aerial viewpoints, enabling wide-ranging scene sampling with drones. Specifically, AirSim360 focuses on three key aspects: a render-aligned data and labeling paradigm for pixel-level geometric, semantic, and instance-level understanding; an interactive pedestrian-aware system for modeling human behavior; and an automated trajectory generation paradigm to support navigation tasks. Furthermore, we collect more than 60K panoramic samples and conduct extensive experiments across various tasks to demonstrate the effectiveness of our simulator. Unlike existing simulators, our work is the first to systematically model the 4D real world under an omnidirectional setting. The entire platform, including the toolkit, plugins, and collected datasets, will be made publicly available at \href{https://insta360-research-team.github.io/AirSim360-website/}{https://insta360-research-team.github.io/AirSim360-website/}.}
\end{abstract}

\begin{table*}[t!]
\centering
\fontsize{8.5}{8.5}\selectfont
\renewcommand{\arraystretch}{1.5}
\caption{Comparison of Simulators. This table compares recent simulation platforms across nine key dimensions, including rendering quality, interface support, and data generation flexibility, highlighting the advantages of our generative-AI-oriented simulator. Our \colorbox{gray!25}{\textbf{specific characters}} are highlighted. ``Ent'' indicate the entity segmentation that cover the whole image.}
\label{tab:tabel1}
\resizebox{\textwidth}{!}{
\setlength{\tabcolsep}{1.5mm}
\begin{tabular}{lccccccc}
\toprule
\textbf{Platform} & \textbf{AirSim~\citep{shah2017airsim}} & \textbf{CARLA~\citep{dosovitskiy2017carla}} & \textbf{Cosys-AirSim~\cite{jansen2023cosys}} & \textbf{OmniGibson~\citep{li2024behavior1k}} & \textbf{UnrealZoo~\cite{ zhong2025unrealzoo}} & \textbf{OpenFly~\citep{gao2025openfly}} & \textbf{AirSim360 (ours)} \\
\midrule
\textbf{\textit{Realism level}} & Medium & High & High & Medium & High & High & High \\
\textbf{\textit{Configurable Dynamics}} & No & No & No & No & No & No & \colorbox{gray!25}{Yes} \\
\textbf{\textit{API type}} & Python & Python & Python & Python & Python & Python & \colorbox{gray!25}{Python / Blueprint} \\
\textbf{\textit{Scenario type}} & Outdoor/Indoor & Outdoor & Outdoor/Indoor & Outdoor/Indoor & Outdoor/Indoor & Outdoor & Outdoor/Indoor \\
\textbf{\textit{UAV simulator}} & Yes & No & Yes & No & Yes & Yes & Yes \\
\textbf{\textit{Annotation capability}} & Dep/Seg & Dep/Seg/Ins & Dep/Seg/Ins & Dep/Seg/Ins & Dep/Seg/Ins & Dep/Seg/Ins & \colorbox{gray!25}{Dep/Seg/Ent} \\
\textbf{\textit{Custom label support}} & No & No & Yes & Yes & No & No & \colorbox{gray!25}{Yes} \\
\textbf{\textit{Panoramic image}} & Yes & No & No & No & No & No & \colorbox{gray!25}{Yes} \\
\textbf{\textit{Rendering engine}} & UE 4.27 & UE 5 & UE 5.2 & IsaacSim & UE 5 & UE 5 & \colorbox{gray!25}{UE 4.27 - UE 5.6} \\
\bottomrule
\end{tabular}}
\end{table*}

\section{Introduction}


Embodied intelligence~\citep{cangelosi2015embodied} with \ql{omnidirectional perception}~\citep{shi2017design,lin2025one} has gained increasing attention \ql{due to the 360\textdegree{} full view in spatial intelligence. It can benefit various robotic applications~\citep{collins2021review}, such as omnidirectional obstacle avoidance~\citep{zavlangas2000fuzzy} during navigation tasks~\citep{qi2019amodal}.}


\ql{Different from large-scale perspective image datasets~\citep{deng2009imagenet, lin2014microsoft}, omnidirectional data~\citep{sekkat2020omniscape, coors2018spherenet} remain scarce~\citep{feng2025dit360} due to the limited usage of 360° cameras in daily life, not to mention the exhaustive human labeling required for many tasks. As a result, most panoramic methods are restricted by small datasets and have scarcely explored data scaling~\citep{ahsan2021effect}.}

\ql{Inspired by recent advances in simulation platforms~\citep{sobchyshak2025pushing,shah2017airsim,dosovitskiy2017carla,jansen2023cosys,li2024behavior1k,zhong2025unrealzoo,gao2025openfly}, a straightforward solution is to rotate agent across multiple angles in simulator to capture an omnidirectional view, including both images and corresponding ground truth. However, this approach introduces two major issues. First, it is computationally inefficient, requiring repeated rendering and significantly increasing data collection time. Second, the definitions of ground-truth signals are not aligned with those in the perspective domain. For example, omnidirectional depth~\citep{li20252} represents slant range along the viewing ray rather than the orthogonal z-axis distance used in perspective projection.}



%

\ql{In this work}, we focus on building AirSim360, an omnidirectional simulation platform in the drone view to model the 4D real world that consists of high-quality static environments and movable pedestrians. \ql{The reason we choose UAV as our agent because it can sample much more data by exploring a wider range of spaces than a ground-based one.} \ql{Based on the Unreal Engine (UE) 5 series for scene rendering, AirSim360 integrates custom dynamics and communication modules into UE, enabling UAVs to execute actions driven by an external physical modeling engine and serving as the core engine for our entire data-collection toolkit.} \yuhang{Table~\ref{tab:tabel1} summarizes the core capabilities of Airsim360, which offers a comprehensive API suite ranging from data acquisition (top) to flight control interfaces (bottom). Unlike existing platforms, our simulator supports full runtime interaction, enabling the generation of video-level panoptic segmentation annotations.}

\ql{Specifically, our AirSim360 has three main characters including render-aligned data and label generation, interactive pedestrian-aware system, and automated trajectory generation paradigm.}
\ql{For data collection, we} adopt the equirectangular projection (ERP) as an omnidirectional representation, \ql{which can compress multiple perspective views in a single continuous image}. However, this representation introduces several challenges for both image content and human- or pixel-level annotations that should be carefully addressed.
On the one hand, \ql{we propose a GPU-side texture copying mechanism based on Render Hardware Interface (RHI) to enable fast and seamless stitching of six cube-face views captured simultaneously, while maintaining lighting consistency across all faces.} On the other hand, the depth information is re-calculated to the slant range along the viewing ray and segmentation annotations have semantic and entity-level labels across frames, ensuring complete panoramic coverage and consistent entity identities over time. 
\ql{Furthermore, we develop point-to-point path planning schemes for automatic data collection and introduce an interactive pedestrian-aware system that simulates movable pedestrians with various actions and annotated keypoints, thus enhancing human-centric perception and analysis.}

\ql{Finally, extensive experiments across five panoramic tasks, including depth estimation, semantic/entity segmentation, human keypoint detection, and vision-language navigation, demonstrate that our simulated data transfers effectively to real-world scenarios.}





Our contributions are fivefold.

\begin{itemize}[noitemsep, leftmargin=*]
    \item We propose a simulation platform, Airsim360, with native support for 360-degree aerial scenarios. Built on the UE 5 Series with an aerial dynamics module for high-quality scene rendering, the platform provides offline data generation tools capable of capturing panoramic images and corresponding ground truth, including semantic segmentation, entity segmentation, depth estimation, and 3D human keypoints.
    \item For efficient data collection, we develop point-to-point path planning schemes that enable both automatic sampling and user-designed flight paths. Moreover, we introduce an interactive pedestrian-aware system that synthesizes pedestrian behaviors with detailed annotations and models realistic interactions.
    \item We test our simulated data for various perception and navigation tasks. Extensive experiments demonstrate our data significantly improve performance and robustness when evaluated on real-world validation sets.
    \item To our knowledge, \ql{AirSim360} is the first platform to support omnidirectional navigation for UAVs. Compared with conventional monocular systems, panoramic UAVs provide superior perceptual coverage, enabling more efficient target search with minimal additional motion cost. 
    \item \ql{AirSim360} has excellent backward compatibility with ranging from latest UE 5.6 down to 4.27, our toolkit and benchmark tasks can run across these versions.
\end{itemize}

\section{Related Work}

\paragraph{UAV Dataset}
UAV datasets typically contain raw images and task-related ground truth, such as trajectory waypoints, depth maps, semantic labels, and point clouds. In general, such data are obtained through acquisition of real-world flights with manual annotation~\cite{lee2024citynav, fan2023aerial, wang2024towards} or all-in-one simulation platforms~\cite{misra2018mapping, liu2023aerialvln, airsim2017fsr}. However, existing datasets are designed primarily for perspective views with limited fields of view. In contrast, we focus on an omnidirectional setting with $360^{\circ}$ coverage in aerial scenarios, which enables panoramic UAV tasks such as omnidirectional obstacle avoidance.


\paragraph{Embodied Simulator}
Embodied simulators~\cite{puig2023habitat, chang2017matterport3d, anguelov2010google} are crucial in robotics, such as autonomous driving and robotic grasping. It is because real-world data collection is often challenging, especially for rare long-tail scenarios. As UAVs play an increasingly important role, a range of UAV-oriented simulation platforms have emerged, including AirSim, UnrealCV~\citep{qiu2017unrealcv}, OpenFly, UAVScenes~\citep{wang2025uavscenes} and UnrealZoo. Among them, AirSim is an earlier simulator that utilizes UE for aerial scene rendering, but its latest officially supported version remains UE 4.27. UnrealCV and UnrealZoo are plugins built on top of UE that provide socket-based interfaces to capture RGB images, depth estimation, and semantic segmentation. However, they only support conventional perspective views and provide limited geometric and semantic information at relatively low frame rates, while lacking entity-level discrimination and making them insufficient for complex panoramic UAV tasks. In Table~\ref{tab:tabel1}, our UAV360 platform additionally enables omnidirectional perception with full 360-degree coverage, and offers entity-level segmentation and 3D keypoint annotations to support advanced panoramic UAV applications at high frame rate.

\paragraph{Panoramic Task}
Panoramic visual tasks in understanding~\cite{xu2022pandora, yang2019pass} and generation~\cite{chang2018generating, chen2022text2light, hold2019deep} have rapidly advanced in support of spatial intelligence~\cite{lin2025flightgapsurveyperspective}, where equirectangular projection (ERP) remains the most prevalent representation due to its straightforward mapping from spherical coordinates to a rectangular image. However, existing methods typically rely on real-world datasets with limited scale and diversity~\cite{launet, chou2020360, yang2021capturing, xu2018predicting, pintore2021deep3dlayout, armeni2017joint}, leaving their generalization ability across domains underexplored. In this work, we systematically benchmark depth estimation, pixel-level scene parsing, human keypoint detection, and UAV navigation under both in-domain and out-of-domain settings, demonstrating the merit of the proposed UAV360 platform.

\begin{figure*}[t!]
 \centering
\includegraphics[width=0.99  \linewidth]{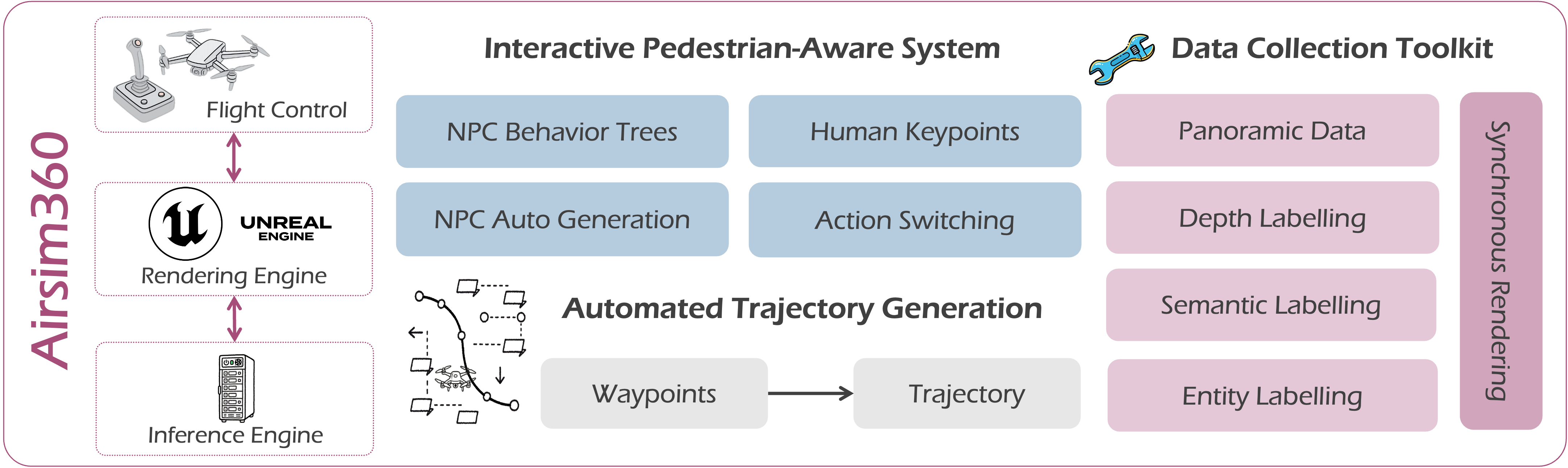}

  \caption{The left panel depicts the core interaction architecture of Airsim360, comprising the flight control module, rendering engine, and inference engine. The middle and right panels showcase three purpose-built data generation modules: the Interactive Pedestrian-Aware System, Automated Trajectory Generator, and Data Collection Toolkit. }
 \label{fig:framework}
\end{figure*}
\section{AirSim360 Platform}
\label{sec:platform}
\ql{As shown in Figure~\ref{fig:framework}, AirSim360 is a large-scale omnidirectional simulation platform for drone scenarios. Unlike other platforms such as AirSim and UnrealZoo, we emphasize consistency and compatibility among three key components in low-altitude ground environments, including static surroundings, the UAV, and human actors, which aligns closely with the main characteristics of the human-centric real world.}

\ql{In the following subsections, we first introduce our simulation architecture and its core design, including the simulation environment, flight control, and communication system for efficient panoramic data sampling. Next, we describe the off-line data collection system, which generates panoramic images and pixel-level ground truth such as semantic segmentation, entity segmentation, and human keypoints using an automated trajectory generation tool. Finally, we present the interactive pedestrian-aware system, which enables realistic human behaviors and interactions within the simulation environment. }

\subsection{Overview}
\label{sec:architecture}
\ql{Built upon high-quality rendering capabilities of Unreal Engine, AirSim360 is an online closed-loop simulator compatible with multiple versions ranging from UE 4.27 to UE 5.6. By default, we adopt UE5 as our setting, considering its significant improvements in dynamic illumination, geometric detail, and overall scalability.}

\ql{To improve user accessibility, we further optimize the communication module between virtual sensors and the UAV flight controller while providing an open external interface such as Vision-Language-Action (VLA)~\cite{jiang2025survey} for flexible integration. In the current system, the external model can return high-level control commands or a direct target position. These values are then parsed by our custom flight control module into the corresponding thrust and torque for each of the four rotors.}

\ql{Compared with other platforms such as AirSim and UnrealZoo, we independently compile a flight control module and integrate it deeply into UE. This design allows us to simulate various types of drone through a simple user prompt that can enhance the generalization and flexibility of our system.} \ql{More details of the implementation are provided in the appendix, as they fall beyond our primary focus on omnidirectional simulation.}

\subsection{Data Collection Toolkit}
\label{sec:offline_tools}
\ql{Apart from} online simulation, AirSim360 provides a powerful toolkit for offline data collection to overcome \ql{scale limitations} of existing omnidirectional datasets, \ql{particularly in pixel-level annotations that are closely related to semantic and geometric understanding.} 


\begin{figure*}[htbp]
 \centering
\includegraphics[width=0.99\linewidth]{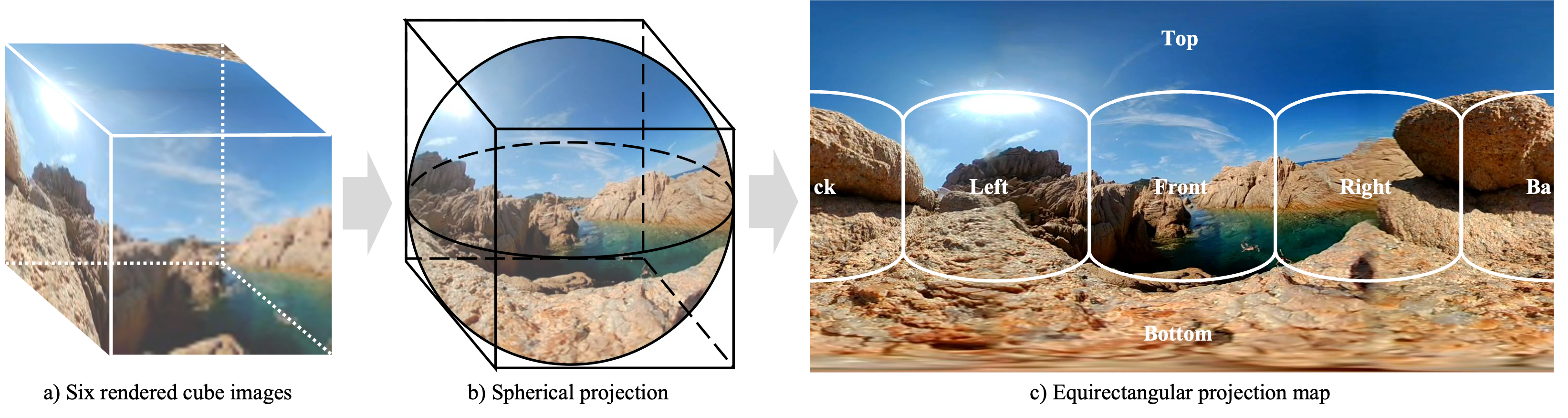}

  \caption{The equirectangular projection serves as a standard representation for panoramic imagery. In our implementation, we configured six identically calibrated cameras facing different directions. Since these are idealized pinhole cameras without lens distortion in the simulation environment, their outputs can be seamlessly stitched into a corresponding panoramic image through camera stitching algorithms.}
  \vspace{-5mm}
 \label{fig:pro_cube_erp}
\end{figure*}

\subsubsection{Render-Aligned Data and Label Generation}
\label{sec:render_aligned}
\ql{Inspired by the stitching process of omnidirectional cameras, our data, whether the original images or their corresponding ground truth, are generated by stitching six cube-face views together. Each view corresponds to one of the six directions, namely front, back, left, right, up, and down, with a 90-degree field of view. Specifically, we obtain six non-overlapped cube images $I^{6\times H_c \times W_c}_{c}$ as input and then produce an equirectangular projection map (ERP) $I^{H_e \times W_e}_{e}$. Here, $H_c$, $W_c$, $H_e$ and $W_e$ are the height and width of the cube and erp images, respectively. For clear illustration of such a process, we use the spherical projection which serves as a bridge between the input and output representations in Figure~\ref{fig:pro_cube_erp}. The detailed equation is in our appendix. }

\ql{In the following, we introduce the unique challenges posed by our stitching process for each data type, along with the corresponding solutions.}

%
%
%

\noindent \textbf{Panoramic Image:} \ql{Capturing multimodal data simultaneously from six directions significantly increases GPU rendering load and storage pressure, resulting in substantial frame rate drops and difficulty in maintaining high throughput at high resolutions. For multi-view image rendering, we have redesigned the composition mechanism for six camera-associated images from ground up. After establishing corresponding camera rendering views, we utilize internal functions of Render Hardware Interface (RHI) to perform GPU-side texture resource copying, enabling one-time stitching of the six images. This approach thus avoids the drawbacks of secondary stitching within material nodes via blueprints.}



\noindent \textbf{Depth Information}: \ql{In a perspective image, depth is defined as the distance between a 3D point and the camera center along the optical axis. This definition differs in the omnidirectional view, where no tangent plane exists. Therefore, in our setting, depth is defined as the distance from the center of the camera to the point along the viewing ray.}

\ql{Since Z-Depth can be directly acquired from the precomputed depth buffer (Z-Buffer) in the Unreal Engine, we proposed a material-based pipeline to extract depth data and render the results into a Render Target. This approach allows writing the depth information directly into the alpha channel of the corresponding image. During the actual rendering process, with known camera world coordinates and corresponding intrinsic and extrinsic parameters, the precise distance from the camera to each spatial point can be computationally determined.}

\noindent \textbf{Semantic Segmentation}: \ql{We assign a semantic label to each pixel to enhance scene understanding. In our implementation within Unreal Engine, we represent a set of semantic categories by assigning specific RGB values. To achieve this, we utilize the Stencil Buffer from the graphics rendering pipeline to assign colors to the static mesh actuator. The Stencil Buffer stores an integer value between 0 and 255 for each pixel in the scene. Through custom-designed post-process material, these values are transformed into designated color outputs, facilitating the acquisition of semantic labels.}

\noindent \textbf{Entity Segmentation}: \ql{Unlike other simulators that focus only on partial instances, we consider all entities present in the scene, ensuring complete coverage across the entire image. However, this design presents a challenge related to quantity limitations. The Stencil Buffer mechanism restricts the total number of assignable categories to 256, which is insufficient for the large number of objects typically involved in entity segmentation tasks, particularly in complex scenarios. To overcome this limitation, we develop a dedicated entity segmentation method capable of labeling all static mesh actors, skeleton mesh actors, and landscape elements within the scene. This approach allows us to obtain complete and fine-grained segmentation results for any given frame.} 

\noindent \textbf{Synchronous Rendering among Various Sensors}: \ql{Various sensors should capture data synchronously without latency. Considering that UAV-mounted cameras are in constant motion, we have deactivated the \textit{Capture Every Frame} option for all rendering cameras, significantly reducing GPU resource consumption. Since we have custom-designed the cameras and various image sensors, we introduced an Event Dispatcher to synchronize all sensors with a unified trigger signal, enabling simultaneous acquisition of multiple data types.}

\begin{figure*}[t!]
  \centering
\includegraphics[width=0.99  \linewidth]{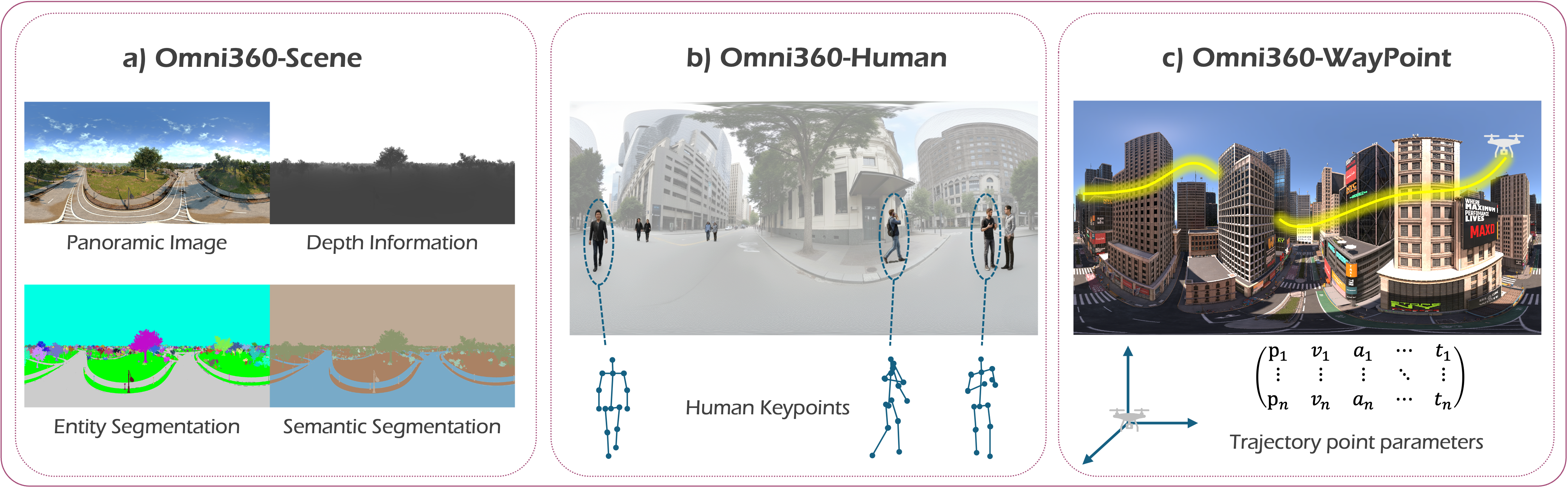}
  \caption{Visualization of Omni360-X. a) Visualizes Render-Aligned Data with corresponding labels (\cref{sec:render_aligned}). b) Demonstrates pedestrian-aware label generation under panoramic views (\cref{sec:ipais}). c) Shows trajectory synthesis from sparse waypoints using the Minimum Snap (\cref{sec:traj_synthesis}).}
  \vspace{-4mm}
  \label{fig:vis_omni360-x}
\end{figure*}

\subsubsection{Interactive Pedestrian-Aware System}
\label{sec:ipais}

\begin{figure}[t]
  \centering
    \includegraphics[width=0.99\linewidth]{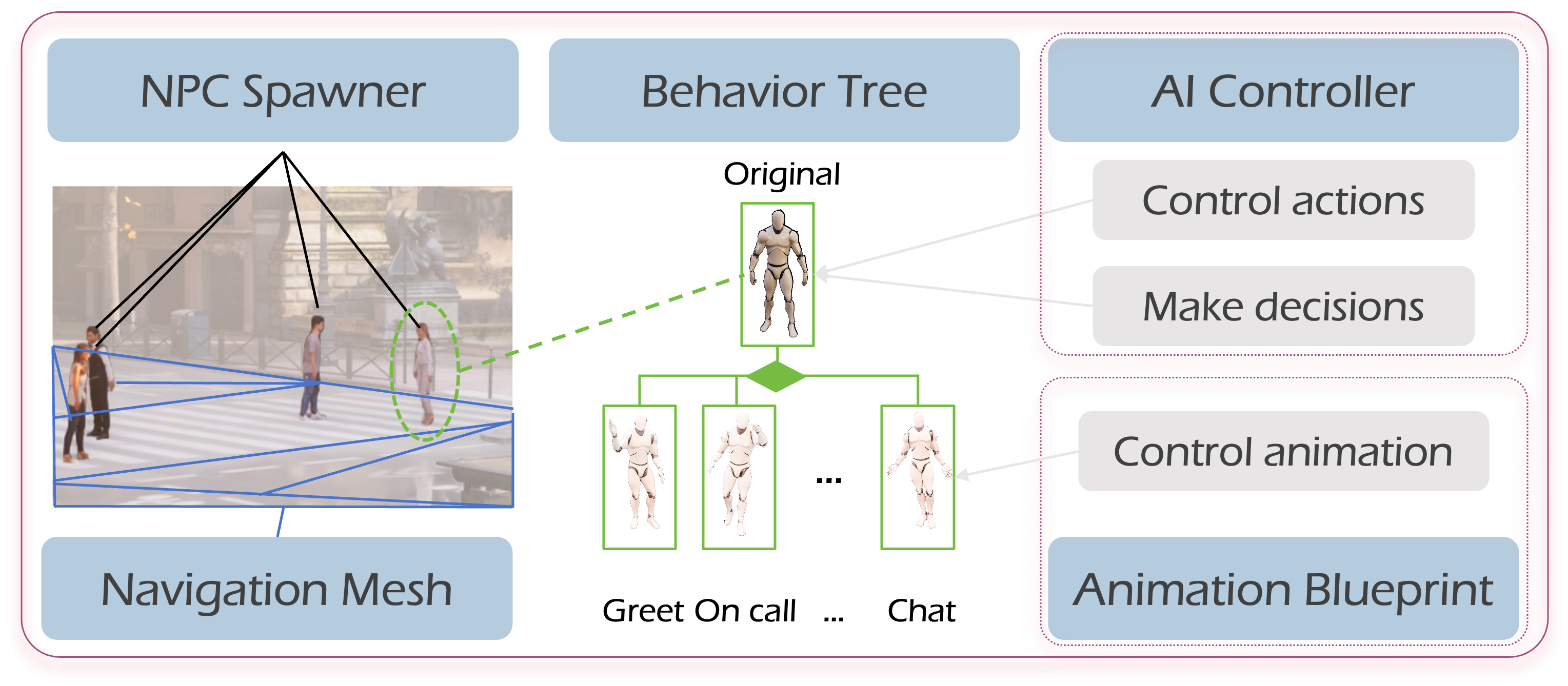}
  \caption{Conceptual diagram of the IPAS logic, showing the interplay between the Behavior Tree, State Machine, and the message-passing system for autonomous agent interaction.}
  \vspace{-6mm}
  \label{fig:ipais_logic}
\end{figure}

\ql{Scenarios involving humans are critically important in low-altitude real-world environments. Therefore, simulating realistic human behavior is a key aspect of our platform. We begin it from three perspectives. First, we allow users to create a customizable number of pedestrians within a defined active area, automatically assigning various behaviors to them. Second, we enable autonomous interactions by combining NPC Behavior Trees with State Machines. For example, a pedestrian state machine can trigger transitions based on a multi-actor message dispatch and receive mechanism, allowing an agent to switch from a walking state to a chatting state when meeting another agent, or to randomly activate an OnPhoneCall state while walking (see Figure~\ref{fig:ipais_logic}). Third, human keypoints are generated in real time as pedestrians interact, ensuring temporally consistent annotations for downstream perception tasks.}

\ql{Generating pedestrian keypoint data presents two main challenges. First, a framework capable of supporting autonomous pedestrian motion is required to accommodate diversity in body movements. Second, a method for binding identical skeletal keypoints across different characters is essential to avoid positional inaccuracies introduced by manual keypoint annotation.}

\ql{To address these issues, we design Interactive Pedestrian-aware System (IPAS), a system that enables autonomous and interactive movements among pedestrians. Additionally, we implement a Blueprint-based approach that invokes pre-existing universal skeletal points in the Skeletal Mesh via blueprint functions. For keypoints not included in the standard skeletal framework, we use the \textit{Add Socke} method to incorporate them into the Skeleton Tree, thereby enabling users to access and output all designed body keypoint coordinates.}


    


\subsubsection{Automated Trajectory Generation Paradigm}
\label{sec:traj_synthesis}
The ground truth trajectories in previous platforms \ql{have largely relied on human control of the drone.} To make the sampling process more efficient, we adopt Minimum Snap trajectory planning~\cite{mellinger2011minimum} into our simulation pipeline. A collector only needs to specify a few key waypoints within the scene, after which the system automatically generates a smooth and realistic trajectory that adheres to the dynamic constraints of the UAV. By adjusting parameters such as maximum velocity and acceleration, the resulting polynomial coefficients of the trajectories can be applied directly to real quadrotors.

\section{Datasets Collection}

\ql{Based on our proposed AirSim360 simulator, we collect several large-scale datasets for omnidirectional scene understanding. Then, we introduce the datasets gathered from a variety of Unreal Engine 5 (UE5) scenarios. Specifically, Omni360-X is a large-scale panoramic dataset containing 60K nonduplicated frames filtered using SSCD~\citep{pizzi2022self}. To serve diverse research objectives, Omni360-X is organized into three subsets, including Scene, Pedestrian, and Trajectory, where each emphasizes a specific aspect of 360-degree understanding. The visualization is shown in Figure~\ref{fig:vis_omni360-x}.}
\begin{table}[t!]
\centering
\normalsize 
\caption{Composition of the Omni360-Scene Dataset. The dataset includes four Unreal Engine scenes, each depicting a unique urban environment. For scenes of varying scales, we collected panoramic data with corresponding labels, while the final column reports the total number of labels per scene.}
\label{tab:Omni360-Scene Dataset}
\setlength{\tabcolsep}{10pt} 
\renewcommand{\arraystretch}{1.6} 
\resizebox{\columnwidth}{!}{%
\begin{tabular}{l r r c c}
\toprule
\textbf{Scenarios Name} & \textbf{Area (m$^2$)} & \textbf{Nums} & \textbf{Label types} & \textbf{Sem. Cat.} \\
\midrule
\textbf{\textit{City Park}} & 800,000 & 25,600 & Dep/Seg/Ins & 25 \\
\textbf{\textit{Downtown West}} & 60,000 & 6,800 & Dep/Seg/Ins & 29 \\
\textbf{\textit{SF City}} & 250,000 & 22,000 & Dep/Seg/Ins & 20 \\
\textbf{\textit{New York City}} & 44,800 & 6,600 & Dep/Seg/Ins & 25 \\
\bottomrule
\end{tabular}%
}
\vspace{-6mm}
\end{table}

\ql{More detailed information about the Omni360-X dataset is provided in the supplementary material due to page limitations.}

\noindent \paragraph{Omni360-Scene}
\ql{We focus primarily on scene parsing within Omni360-Scene. Inspired by ADE20K~\cite{zhou2019semantic}, Omni360-Scene contains more than 60K images annotated with depth and panoptic segmentation. Panoptic segmentation has two components: semantic and entity segmentation. For semantic labeling, we first extract semantic nouns from UE 5, which can be directly used for understanding open-world or open-vocabulary scenes. We then design a hierarchical semantic tree to ensure semantic consistency across different scenes. or entity segmentation, we further decompose stuff categories, such as trees and buildings, into individual entities. This is made possible by the simulator, which enables detailed entity separation that would be extremely challenging to achieve manually.}

\gx{In Table~\ref{tab:Omni360-Scene Dataset}, we present the data statistics for the four scenarios within Omni360-Scene. The dataset contains a total of 61,000 images captured from four scenarios, which demonstrate significant diversity in both physical scale (from 44,800 m² to 800,000 m²) and semantic complexity (20 to 29 categories). All scenarios provide comprehensive ground truth for Depth, Semantic Segmentation, and Instance Segmentation}

\noindent \paragraph*{Omni360-Human}
\label{sec:Omni360_Human}
\ql{To better understand pedestrian behavior, we adopt 3D monocular human localization as our primary task, as it measures both human positions and postures in the 3D world, which are closely related to behavioral understanding. Accordingly, Omni360-Human contains approximately 100K samples covering more than 10 pedestrian behaviors across diverse camera distances and viewpoints in about 6 scenes. } 

\ql{As shown in Figure~\ref{fig:vis_omni360-x} (b), each sample contains both camera information, including the camera’s absolute position and rotation angles in the world coordinate system, and pedestrian information, such as the locations and rotation angles of human keypoints. }

\noindent \paragraph{Omni360-WayPoint}
\ql{Building upon our proposed automatic trajectory synthesis scheme, we will release Omni360-WayPoint, an open dataset containing over 100,000 UAV waypoints, as described in Table~\ref{tab:Omni360-WayPoint}. The trajectories adhere to realistic flight dynamics and include route variants parameterized by different maximum level-flight speeds, enabling analyses across diverse kinematic regimes.} 
\begin{table}[t!]
\centering
\small
\caption{Summary of the Omni360-WayPoint dataset. Flight paths of different lengths and sampling rates were generated across four outdoor scenes with physically consistent UAV dynamics.}
\label{tab:Omni360-WayPoint}
\setlength{\tabcolsep}{4pt} 
\renewcommand{\arraystretch}{1.5} 
\resizebox{\columnwidth}{!}{%
\begin{tabular}{@{}>{\centering\arraybackslash}m{1.4cm}
                 >{\centering\arraybackslash}m{2.2cm}
                 >{\centering\arraybackslash}m{1.4cm}
                 >{\centering\arraybackslash}m{1.3cm}
                 >{\centering\arraybackslash}m{1.1cm}
                 >{\centering\arraybackslash}m{1.5cm}@{}}
\toprule
\textbf{Thumbnail} &
\textbf{Scenario} &
\textbf{Scene type} &
\textbf{Count} &
\textbf{Spacing} &
\textbf{Length range}\\
\midrule
\raisebox{0.2cm}{\adjustimage{height=0.65cm,valign=m}{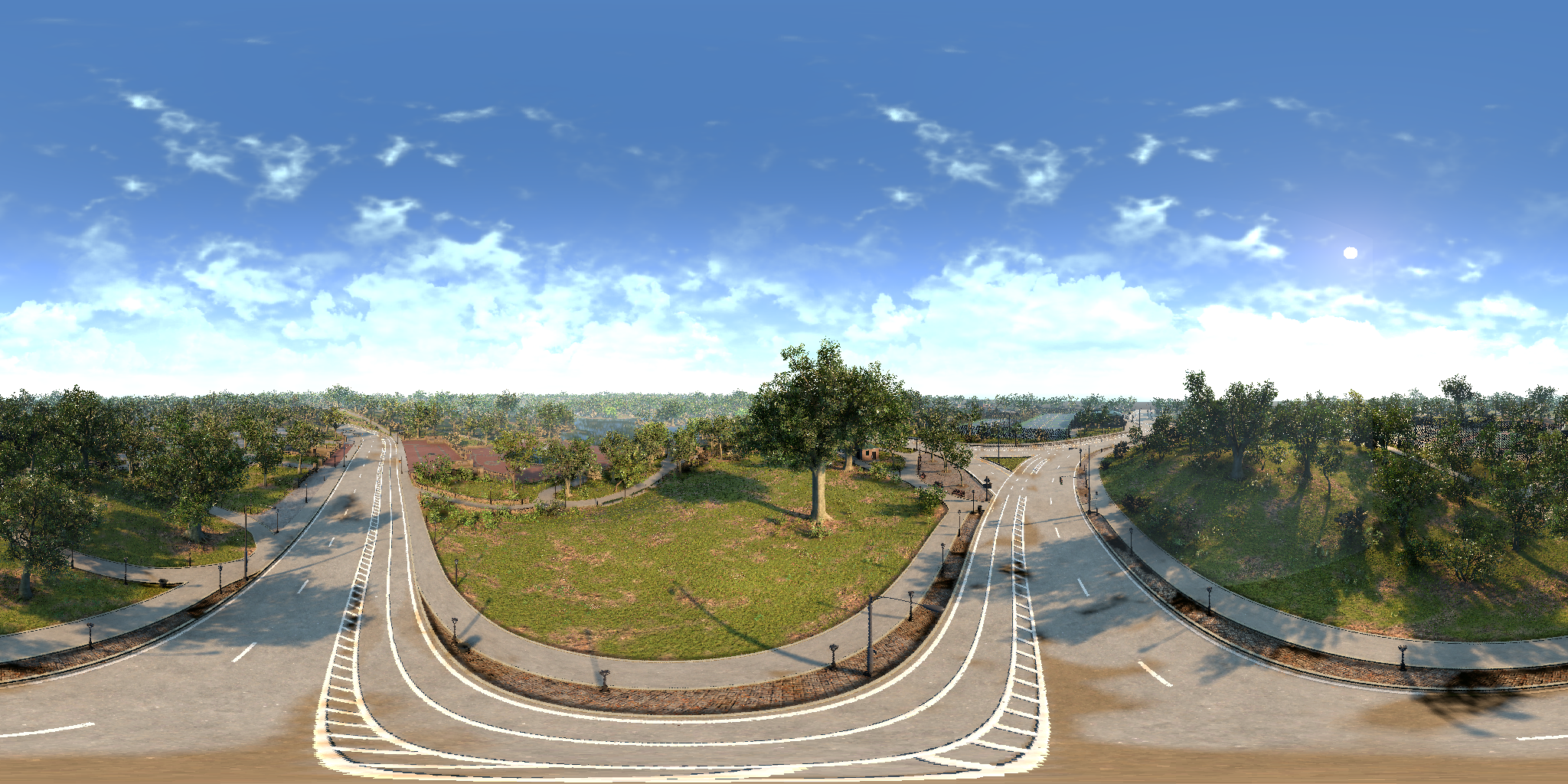}} & \textbf{\textit{City Park}}     & outdoor & 20\,000 & 0.5 & [50,\,150]\\
\raisebox{0.2cm}{\adjustimage{height=0.65cm,valign=m}{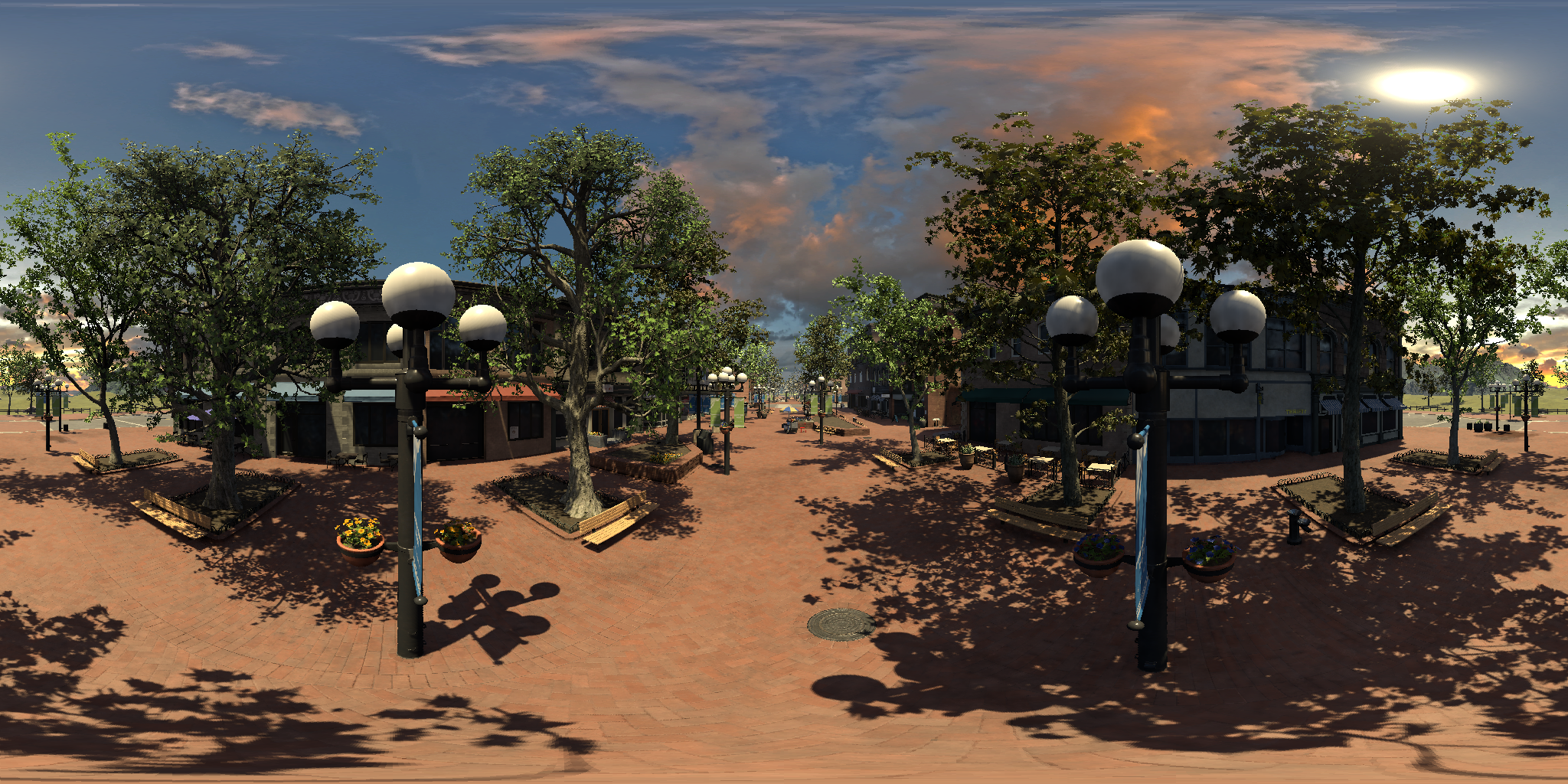}} & \textbf{\textit{Downtown West}} & outdoor &  5\,000 & 0.2 & [20,\,50]\\
\raisebox{0.2cm}{\adjustimage{height=0.65cm,valign=m}{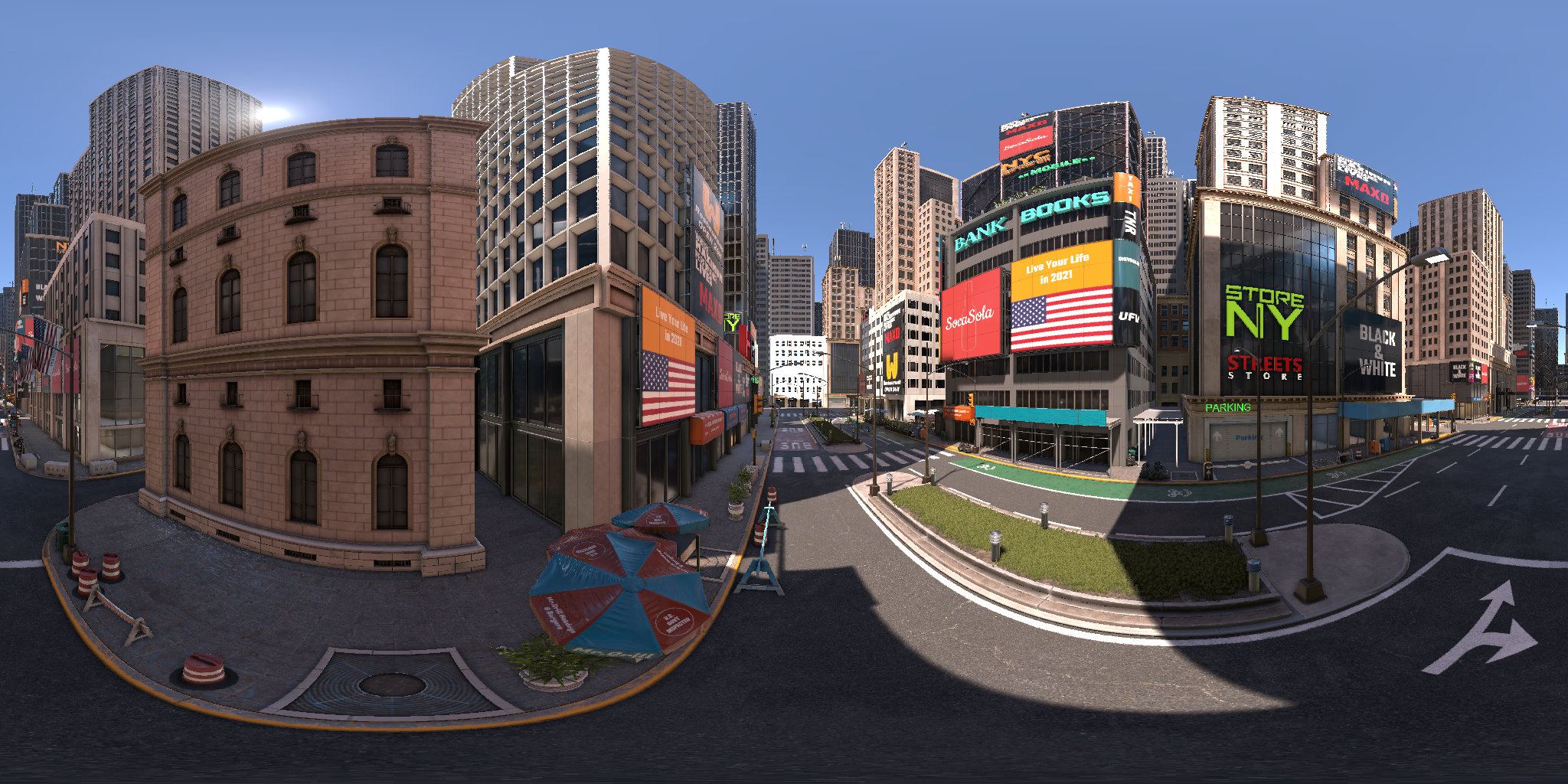}} & \textbf{\textit{New York City}} & outdoor &  5\,000 & 0.2 & [20,\,50]\\
\raisebox{0.2cm}{\adjustimage{height=0.65cm,valign=m}{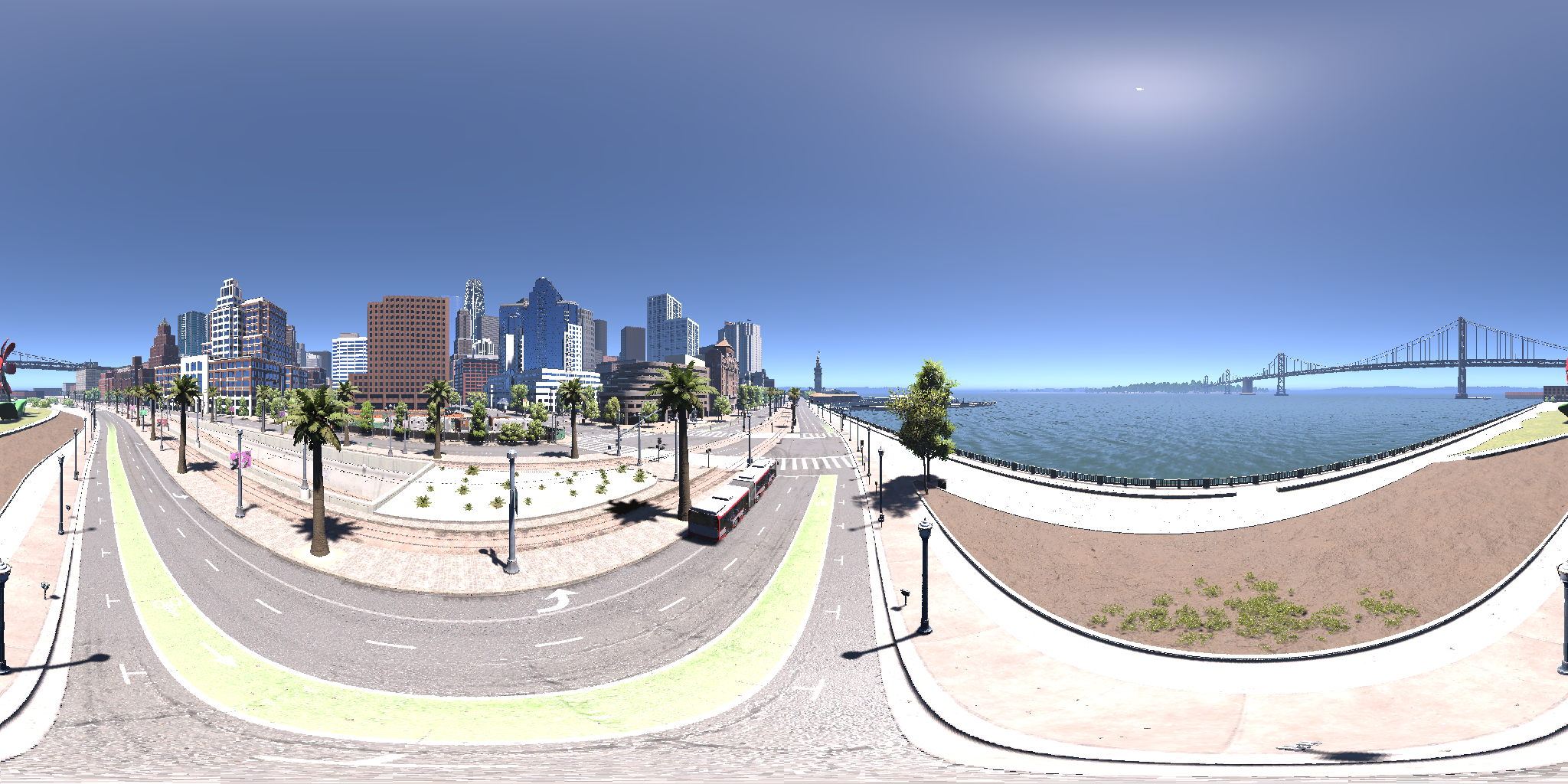}} & \textbf{\textit{SF City}}       & outdoor & 20\,000 & 0.5 & [50,\,150]\\
\bottomrule
\end{tabular}%
}
\end{table}
\ql{Such datasets can serve as various targets. For example, they can provide physics-consistent supervision for perception and state estimation, and enable trajectory prediction and system identification for model-based control and reinforcement learning. Moreover, they align instruction–video–action for VLA training, strengthen 3D reconstruction and mapping with accurate poses.}

\section{Experiments}
\ql{In this section, we conduct several ablation studies on the AirSim360 platform to demonstrate the effectiveness of our design choices. We then present and benchmark the sampled dataset in multiple tasks, indicating the benefits of our proposed simulator for real-world scenarios.}

\subsection{Ablation Study on AirSim360}

\yuhang{We conduct several comparison experiments on a workstation equipped with an NVIDIA RTX 4060 Ti (8 GB), an Intel i7-14700, and 32 GB of RAM. The evaluation covers two operational modes: perspective camera rendering and panoramic image stitching and transmission. Across all settings, our system consistently achieves higher data throughput and improved frame rates than conventional approaches.}

\begin{table}[t!]
\centering
\footnotesize
\caption{Performance metrics for frame capture settings.}
\label{tab:fps_performance}
\begin{tabular}{@{}lcc@{}}
\toprule
\textbf{Capture Every Frame} & \textbf{FPS} & \textbf{GPU Time} \\ \midrule
Enable & 20 & 54 ms \\
Disable & 29 & 35 ms \\ \bottomrule
\end{tabular}
\end{table}
\yuhang{
Each camera in the rendering engine is processed as an independent rendering pass. Because the cameras move in real time, we optimize their scene-scanning pattern to improve efficiency. As shown in Table~\ref{tab:fps_performance}, this optimized scanning strategy yields a 45\% increase in per-camera frame rate.}
\begin{table}[t!]
\footnotesize
\centering
\caption{Comparison of frame rates across different platforms with 6 cameras.}
\label{tab:platform_rhi_fps}
\begin{tabular}{@{}lcc@{}}
\toprule
\textbf{Platform} & \textbf{Nums of Camera} & \textbf{FPS} \\ \midrule
Not Optimized & 6 & 14 \\
Airsim360 & 6 & 18 \\ \bottomrule
\end{tabular}
\end{table}

\yuhang{
Given the panoramic design of our platform, we restructure the composition and data-transmission pipeline for the six camera feeds in the rendering engine. Leveraging the Render Hardware Interface (RHI) described in Section~\ref{sec:render_aligned}, we reimplement the C++ pipeline, increasing the frame rate from 14 to 18 FPS.
}

\subsection{Monocular Pedestrian Distance Estimation}

\yuhang{To assess the impact of synthetic data, we apply IPAS (view in \cref{sec:ipais}) to Monocular Pedestrian Distance Estimation (MPDE) in Table~\ref{tab:MPDE}. The generated Omni360-Human dataset provides 3D keypoint annotations in the world coordinate system. Using diverse simulated scenes for data augmentation, our approach achieves improved results on several MPDE benchmarks. The model is trained with Omni360-Human and nuScenes~\cite{caesar2020nuscenes}, and evaluated on Omni360-Human, nuScenes, KITTI~\cite{geiger2012we}, and FreeMan~\cite{wang2024freeman}, reporting both Euclidean and angular distance errors following MonoLoco++~\cite{bertoni_2021_its}.}

\begin{table}[t!]
\centering
\footnotesize
\caption{The MPDE results trained on two datasets, evaluated across three public benchmarks and one control set, \textit{Omni360}. The \textit{Omni360} refers to the dataset Omni360-Human.}
\label{tab:MPDE}
\renewcommand{\arraystretch}{1.1}
\setlength{\tabcolsep}{2.3pt}

\begin{tabular}{
    >{\centering\arraybackslash}m{1.4cm}
    >{\centering\arraybackslash}m{1.4cm}
    >{\centering\arraybackslash}m{0.9cm}
    >{\centering\arraybackslash}m{0.9cm}
    >{\centering\arraybackslash}m{1.1cm}
    >{\centering\arraybackslash}m{1.1cm}
}
\toprule
\textbf{\makecell{Training\\Set}} &
\textbf{\makecell{Test\\Set}} &
\textbf{\makecell{Dist.\\Err}} &
\textbf{\makecell{Ang.\\Err}} &
\textbf{\makecell{Ang. Err\\(Pub)}} &
\textbf{\makecell{Dist. Err\\(Pub)}} \\
\midrule

\multirow{5}{*}{\centering\textbf{\textit{nuScenes}}} 
 & \textit{nuScenes} & 1.078 & 31.90 & \multirow{3}{*}{\centering 21.21} & \multirow{3}{*}{\centering 0.484} \\
 & \textit{KITTI} & 0.822 & 31.50 &  &  \\
 & \textit{FreeMan} & 0.260 & 17.00 &  &  \\
\cline{2-6} 
\multicolumn{1}{c}{} & \textit{Omni360} & 2.439 & 33.30 &  &  \\

\midrule

\multirow{5}{*}{\centering\makecell{\textbf{\textit{nuScenes}}\\+ \textbf{\textit{Omni360}}}} 
 & \textit{nuScenes} & 1.068 & 30.70 & \multirow{3}{*}{\centering {\textbf{17.02}}} & \multirow{3}{*}{\centering {\textbf{0.458}}} \\
 & \textit{KITTI} & 0.809 & 31.20 &  &  \\
 & \textit{FreeMan} & 0.228 & 11.60 &  &  \\
\cline{2-6}
\multicolumn{1}{c}{} & \textit{Omni360} & 1.779 & 15.20 &  &  \\

\bottomrule
\end{tabular}
\end{table}

\yuhang{As shown in Table~\ref{tab:MPDE}, training with the Omni360-Human dataset consistently improves performance across all three public test sets. The average angular error decrease from 21.21° to 17.02°, and the mean distance error from 0.484 m to 0.458 m. To better reflect real-world camera configurations, synthetic data are generated with a 20° pitch angle. The largest test set, FreeMan, exhibit the most significant improvement, with the angular error reduced from 17° to 11.6°. These results highlight the effectiveness of synthetic data in enhancing monocular pedestrian distance estimation.}

\vspace{-1mm}
\subsection{Panoramic Depth Estimation}
\vspace{-1mm}

To evaluate the generalization and domain adaptability of our Omni360 dataset, we conduct out-of-domain and cross-domain experiments for panoramic depth estimation. All datasets consist of outdoor synthetic data. The UniK3D model~\cite{piccinelli2025unik3duniversalcameramonocular} is used as the baseline and fine-tuned after its official implementation. We compare the results with the Deep360 dataset~\cite{Li_Jin2022MODE}, using Absolute Relative Error(AbsRel), Root Mean Squared Error(RMSE), and a percentage metrics $\delta_1$, where i = 1.25, as evaluation metrics.

\begin{table}[t!]
  \centering
  \caption{
    Quantitative comparison for depth estimation. 
    We fine-tune on the UniK3D model and compare the performance 
    of training on Deep360 versus our Omni360 dataset.
    $\downarrow$ and $\uparrow$ denote lower or higher is better.Best results are in bold.
  }
  \label{tab:depth_ablation}
  \resizebox{\columnwidth}{!}{
  \begin{tabular}{lllccc}
    \toprule
    \textbf{Experiment Type} & \textbf{Training Data} & \textbf{Evaluation Data} & \textbf{AbsRel} $\downarrow$ & \textbf{RMSE} $\downarrow$ & \textbf{$\delta_1$} $\uparrow$ \\
    \midrule
    
    \multirow{2}{*}{Out-of-Domain} 
    & Deep360 & SphereCraft & 8.2570 & 0.0566 & 0.3490 \\
    
    & Omni360 (Ours) & SphereCraft & \textbf{5.4372} & \textbf{0.0435} & \textbf{0.3990} \\
    
    \midrule
    \multirow{2}{*}{Cross-Domain}  
    & Deep360 & Omni360 (Ours) & 0.3600 & 0.0714 & 0.4896 \\
    
    & Omni360 (Ours) & Deep360 & \textbf{0.1762} & \textbf{0.0229} & \textbf{0.6672} \\
    
    \bottomrule
  \end{tabular}
  } 
\end{table}

\noindent \textbf{Out-of-Domain}: Models trained in Deep360 and Omni360 are tested on SphereCraft~\cite{Gava_2024_WACV}. In Table~\ref{tab:depth_ablation}, the model trained on Omni360 achieves better results, showing improved generalization to unseen environments.

\noindent \textbf{Cross-Domain}: We perform cross-domain tests between Deep360 and Omni360. The model trained on Omni360 performs best, indicating stronger robustness and more transferable representations.

\vspace{-1mm}
\subsection{Panoramic Segmentation}
\begin{table}[t!]
\centering
\caption{Results of panoramic segmentation. The mIoU and mAP metrics are used to evaluate performance on the semantic and entity segmentation tasks, respectively.}
\label{Tab:seg}
\footnotesize
\setlength{\tabcolsep}{4pt}
\begin{tabular}{c|cc|c}
\hline
\textbf{Task} & \textbf{WildPASS} & \textbf{Omni360-Scene} & \textbf{Performance}
 \\ \hline
\multirow{2}*{Semantic} & \checkmark & $\circ$ & 58.0 \\
& \checkmark & \checkmark & 67.4 \\ \hline
\multirow{2}*{Entity} & \checkmark & $\circ$ & 24.6 \\
& \checkmark & \checkmark & 38.9 \\ \hline
\end{tabular}
\vspace{-6mm}
\end{table}
In Table~\ref{Tab:seg}, we evaluate semantic and entity segmentation~\citep{qi2022high,qi2022open} in the WildPASS validation set~\cite{yang2021capturing} using OOOPS~\cite{zheng2024open} and Mask2Former~\cite{cheng2022masked}. With the incorporation of our Omni360-Scene segmentation data, both tasks achieve substantial performance gains, highlighting the effectiveness of large-scale data scaling for pixel-level semantic understanding.

\subsection{Benchmark on Omni360-X}
\subsubsection{Panoramic Vision-Language Navigation}
\yuhang{
Vision-Language Navigation (VLN)~\cite{anderson2018vision} for unmanned aerial vehicles (UAVs) represents a challenging yet crucial direction toward embodied intelligence in aerial agents, , requiring joint understanding of spatial scenes and language instructions. Most existing UAV-VLN methods mostly rely on forward-facing cameras, resulting in a narrow field of view and large blind areas that limit target search efficiency and situational awareness.
}

\yuhang{
We introduce a panoramic UAV-VLN task built on our simulation platform, where the UAV perceives the environment through panoramic vision instead of a single forward camera. 
}

\begin{table}[t!]
\centering
\caption{Results of panoramic VLN. Three standard metrics, Success Rate (SR), Success weighted by Path Length (SPL), and Navigation Error (NE), are used to evaluate model performance.}
\label{Tab:vln}
\footnotesize
\setlength{\tabcolsep}{4pt}
\begin{tabular}{cccc}
\hline
\textbf{Model} & \textbf{SR} & \textbf{SPL} & \textbf{NE} \\
\hline
\textit{qwen2.5-vl-72b-instruct} & 0.4 & 0.3843 & 18099.73 \\
\textit{qwen3-vl-plus} & 0.0 & 0.0 & 11436.97 \\
\textit{qwen3-vl-flash} & 0.2 & 0.1945 & 9506.26 \\
\textit{doubao-seed-1-6-251015} & 0.5 & 0.4813 & 10573.89 \\
\hline
\end{tabular}
\end{table}

\yuhang{A set of preliminary experiments is conducted using several vision–language models to assess their ability to interpret panoramic observations for navigation. }

\yuhang{In this section, to demonstrate the You Only Move Once (YOMO) capability of panoramic UAVs in decision-making tasks, we place most targets as salient objects and designed short flight trajectories. Under these conditions, the panoramic UAV can directly move toward the goal without additional yaw rotations or exploration. The similar SPL and SR values reported in Table~\ref{Tab:vln} confirm that panoramic perception enables near-optimal, single-step decision making when the UAV is close to the target, validating the YOMO principle.}

\section{Conclusion}
\ql{To address the issue of lacking large-scale omnidirectional data, we propose AirSim360, a panoramic simulation platform built on Unreal Engine. Inspired by the key elements in 4D real world, AirSim360 focuses on three aspects. First, we introduce a render-aligned data pipeline for generating 360-degree images with pixel-level depth and segmentation annotations in ERP representation. Second, we develop an interactive pedestrian-aware system that enables the study of human behavior through 3D monocular human localization. Third, we present an automated trajectory-generation paradigm that produces realistic aerial trajectories for navigation tasks. Built on top of AirSim360, we collect more than 60K non-duplicate frames. Ablation studies on the UE 5-based design, along with extensive experiments on the collected data, demonstrate the effectiveness of our platform and the benefits of the resulting dataset.}


{
    \small
    \bibliographystyle{ieeenat_fullname}
    \bibliography{main}

\begin{thebibliography}{59}
\providecommand{\natexlab}[1]{#1}
\providecommand{\url}[1]{\texttt{#1}}
\expandafter\ifx\csname urlstyle\endcsname\relax
  \providecommand{\doi}[1]{doi: #1}\else
  \providecommand{\doi}{doi: \begingroup \urlstyle{rm}\Url}\fi

\bibitem[Ahsan et~al.(2021)Ahsan, Mahmud, Saha, Gupta, and Siddique]{ahsan2021effect}
Md~Manjurul Ahsan, MA~Parvez Mahmud, Pritom~Kumar Saha, Kishor~Datta Gupta, and Zahed Siddique.
\newblock Effect of data scaling methods on machine learning algorithms and model performance.
\newblock \emph{Technologies}, 9\penalty0 (3):\penalty0 52, 2021.

\bibitem[Anderson et~al.(2018)Anderson, Wu, Teney, Bruce, Johnson, S{\"u}nderhauf, Reid, Gould, and Van Den~Hengel]{anderson2018vision}
Peter Anderson, Qi Wu, Damien Teney, Jake Bruce, Mark Johnson, Niko S{\"u}nderhauf, Ian Reid, Stephen Gould, and Anton Van Den~Hengel.
\newblock Vision-and-language navigation: Interpreting visually-grounded navigation instructions in real environments.
\newblock In \emph{Proceedings of the IEEE conference on computer vision and pattern recognition}, pages 3674--3683, 2018.

\bibitem[Anguelov et~al.(2010)Anguelov, Dulong, Filip, Frueh, Lafon, Lyon, Ogale, Vincent, and Weaver]{anguelov2010google}
Dragomir Anguelov, Carole Dulong, Daniel Filip, Christian Frueh, St{\'e}phane Lafon, Richard Lyon, Abhijit Ogale, Luc Vincent, and Josh Weaver.
\newblock Google street view: Capturing the world at street level.
\newblock \emph{Computer}, 43\penalty0 (6):\penalty0 32--38, 2010.

\bibitem[Armeni et~al.(2017)Armeni, Sax, Zamir, and Savarese]{armeni2017joint}
Iro Armeni, Sasha Sax, Amir~R Zamir, and Silvio Savarese.
\newblock Joint 2d-3d-semantic data for indoor scene understanding.
\newblock \emph{arXiv preprint arXiv:1702.01105}, 2017.

\bibitem[Bertoni et~al.(2021)Bertoni, Kreiss, and Alahi]{bertoni_2021_its}
Lorenzo Bertoni, Sven Kreiss, and Alexandre Alahi.
\newblock Perceiving humans: from monocular 3d localization to social distancing.
\newblock \emph{IEEE Transactions on Intelligent Transportation Systems}, 2021.

\bibitem[Caesar et~al.(2020)Caesar, Bankiti, Lang, Vora, Liong, Xu, Krishnan, Pan, Baldan, and Beijbom]{caesar2020nuscenes}
Holger Caesar, Varun Bankiti, Alex~H Lang, Sourabh Vora, Venice~Erin Liong, Qiang Xu, Anush Krishnan, Yu Pan, Giancarlo Baldan, and Oscar Beijbom.
\newblock nuscenes: A multimodal dataset for autonomous driving.
\newblock In \emph{Proceedings of the IEEE/CVF conference on computer vision and pattern recognition}, pages 11621--11631, 2020.

\bibitem[Cangelosi et~al.(2015)Cangelosi, Bongard, Fischer, and Nolfi]{cangelosi2015embodied}
Angelo Cangelosi, Josh Bongard, Martin~H Fischer, and Stefano Nolfi.
\newblock Embodied intelligence.
\newblock In \emph{Springer handbook of computational intelligence}, pages 697--714. Springer, 2015.

\bibitem[Chang et~al.(2017)Chang, Dai, Funkhouser, Halber, Niessner, Savva, Song, Zeng, and Zhang]{chang2017matterport3d}
Angel Chang, Angela Dai, Thomas Funkhouser, Maciej Halber, Matthias Niessner, Manolis Savva, Shuran Song, Andy Zeng, and Yinda Zhang.
\newblock Matterport3d: Learning from rgb-d data in indoor environments.
\newblock \emph{arXiv preprint arXiv:1709.06158}, 2017.

\bibitem[Chang et~al.(2018)Chang, Chiu, Chang, Chen, Yao, Lee, and Chu]{chang2018generating}
Shih-Hsiu Chang, Ching-Ya Chiu, Chia-Sheng Chang, Kuo-Wei Chen, Chih-Yuan Yao, Ruen-Rone Lee, and Hung-Kuo Chu.
\newblock Generating 360 outdoor panorama dataset with reliable sun position estimation.
\newblock In \emph{SIGGRAPH Asia}. 2018.

\bibitem[Chen et~al.(2022)Chen, Wang, and Liu]{chen2022text2light}
Zhaoxi Chen, Guangcong Wang, and Ziwei Liu.
\newblock Text2light: Zero-shot text-driven hdr panorama generation.
\newblock \emph{ACM Transactions on Graphics (TOG)}, 41\penalty0 (6):\penalty0 1--16, 2022.

\bibitem[Cheng et~al.(2022)Cheng, Misra, Schwing, Kirillov, and Girdhar]{cheng2022masked}
Bowen Cheng, Ishan Misra, Alexander~G Schwing, Alexander Kirillov, and Rohit Girdhar.
\newblock Masked-attention mask transformer for universal image segmentation.
\newblock In \emph{CVPR}, 2022.

\bibitem[Chou et~al.(2020)Chou, Sun, Chang, Hsu, Sun, and Fu]{chou2020360}
Shih-Han Chou, Cheng Sun, Wen-Yen Chang, Wan-Ting Hsu, Min Sun, and Jianlong Fu.
\newblock 360-indoor: Towards learning real-world objects in 360\textdegree{} indoor equirectangular images.
\newblock In \emph{WACV}, 2020.

\bibitem[Collins et~al.(2021)Collins, Chand, Vanderkop, and Howard]{collins2021review}
Jack Collins, Shelvin Chand, Anthony Vanderkop, and David Howard.
\newblock A review of physics simulators for robotic applications.
\newblock \emph{IEEE Access}, 9:\penalty0 51416--51431, 2021.

\bibitem[Coors et~al.(2018)Coors, Condurache, and Geiger]{coors2018spherenet}
Benjamin Coors, Alexandru~Paul Condurache, and Andreas Geiger.
\newblock Spherenet: Learning spherical representations for detection and classification in omnidirectional images.
\newblock In \emph{Proceedings of the European conference on computer vision (ECCV)}, pages 518--533, 2018.

\bibitem[Deng et~al.(2009)Deng, Dong, Socher, Li, Li, and Fei-Fei]{deng2009imagenet}
Jia Deng, Wei Dong, Richard Socher, Li-Jia Li, Kai Li, and Li Fei-Fei.
\newblock Imagenet: A large-scale hierarchical image database.
\newblock In \emph{2009 IEEE conference on computer vision and pattern recognition}, pages 248--255. Ieee, 2009.

\bibitem[Deng et~al.(2021)Deng, Wang, Xu, Guo, Song, and Yang]{launet}
Xin Deng, Hao Wang, Mai Xu, Yichen Guo, Yuhang Song, and Li Yang.
\newblock Lau-net: Latitude adaptive upscaling network for omnidirectional image super-resolution.
\newblock In \emph{CVPR}, 2021.

\bibitem[Dosovitskiy et~al.(2017)Dosovitskiy, Ros, Codevilla, Lopez, and Koltun]{dosovitskiy2017carla}
Alexey Dosovitskiy, German Ros, Felipe Codevilla, Antonio Lopez, and Vladlen Koltun.
\newblock Carla: An open urban driving simulator.
\newblock In \emph{Conference on robot learning}, 2017.

\bibitem[Fan et~al.(2023)Fan, Chen, Jiang, Zhou, Zhang, and Wang]{fan2023aerial}
Yue Fan, Winson Chen, Tongzhou Jiang, Chun Zhou, Yi Zhang, and Xin Wang.
\newblock Aerial vision-and-dialog navigation.
\newblock In \emph{Findings of the Association for Computational Linguistics: ACL 2023}, pages 3043--3061, 2023.

\bibitem[Feng et~al.(2025)Feng, Zhang, Li, Du, and Qi]{feng2025dit360}
Haoran Feng, Dizhe Zhang, Xiangtai Li, Bo Du, and Lu Qi.
\newblock Dit360: High-fidelity panoramic image generation via hybrid training.
\newblock \emph{arXiv preprint arXiv:2510.11712}, 2025.

\bibitem[Gao et~al.(2025)Gao, Li, You, Liu, Li, Chen, Chen, Tang, Wang, Yang, et~al.]{gao2025openfly}
Yunpeng Gao, Chenhui Li, Zhongrui You, Junli Liu, Zhen Li, Pengan Chen, Qizhi Chen, Zhonghan Tang, Liansheng Wang, Penghui Yang, et~al.
\newblock Openfly: A comprehensive platform for aerial vision-language navigation.
\newblock In \emph{arXiv}, 2025.

\bibitem[Gava et~al.(2024)Gava, Cho, Raue, Palacio, Pagani, and Dengel]{Gava_2024_WACV}
Christiano Gava, Yunmin Cho, Federico Raue, Sebastian Palacio, Alain Pagani, and Andreas Dengel.
\newblock Spherecraft: A dataset for spherical keypoint detection, matching and camera pose estimation.
\newblock In \emph{Proceedings of the IEEE/CVF Winter Conference on Applications of Computer Vision (WACV)}, pages 4408--4417, 2024.

\bibitem[Geiger et~al.(2012)Geiger, Lenz, and Urtasun]{geiger2012we}
Andreas Geiger, Philip Lenz, and Raquel Urtasun.
\newblock Are we ready for autonomous driving? the kitti vision benchmark suite.
\newblock In \emph{2012 IEEE conference on computer vision and pattern recognition}, pages 3354--3361. IEEE, 2012.

\bibitem[Hold-Geoffroy et~al.(2019)Hold-Geoffroy, Athawale, and Lalonde]{hold2019deep}
Yannick Hold-Geoffroy, Akshaya Athawale, and Jean-Fran{\c{c}}ois Lalonde.
\newblock Deep sky modeling for single image outdoor lighting estimation.
\newblock In \emph{Proceedings of the IEEE/CVF conference on computer vision and pattern recognition}, pages 6927--6935, 2019.

\bibitem[Jansen et~al.(2023)Jansen, Verreycken, Schenck, Blanquart, Verhulst, Huebel, and Steckel]{jansen2023cosys}
Wouter Jansen, Erik Verreycken, Anthony Schenck, Jean-Edouard Blanquart, Connor Verhulst, Nico Huebel, and Jan Steckel.
\newblock Cosys-airsim: a real-time simulation framework expanded for complex industrial applications.
\newblock \emph{arXiv preprint arXiv:2303.13381}, 2023.

\bibitem[Jiang et~al.(2025)Jiang, Huang, Qian, Luo, Zhu, Zhong, Tang, Kong, Wang, Jiao, et~al.]{jiang2025survey}
Sicong Jiang, Zilin Huang, Kangan Qian, Ziang Luo, Tianze Zhu, Yang Zhong, Yihong Tang, Menglin Kong, Yunlong Wang, Siwen Jiao, et~al.
\newblock A survey on vision-language-action models for autonomous driving.
\newblock \emph{arXiv preprint arXiv:2506.24044}, 2025.

\bibitem[Lee et~al.(2024)Lee, Miyanishi, Kurita, Sakamoto, Azuma, Matsuo, and Inoue]{lee2024citynav}
Jungdae Lee, Taiki Miyanishi, Shuhei Kurita, Koya Sakamoto, Daichi Azuma, Yutaka Matsuo, and Nakamasa Inoue.
\newblock Citynav: Language-goal aerial navigation dataset with geographic information.
\newblock \emph{arXiv preprint arXiv:2406.14240}, 2024.

\bibitem[Li et~al.(2024)Li, Zhang, Wong, Gokmen, Srivastava, Martín-Martín, Wang, Levine, Ai, Martinez, Yin, Lingelbach, Hwang, Hiranaka, Garlanka, Aydin, Lee, Sun, Anvari, Sharma, Bansal, Hunter, Kim, Lou, Matthews, Villa-Renteria, Tang, Tang, Xia, Li, Savarese, Gweon, Liu, Wu, and Fei-Fei]{li2024behavior1k}
Chengshu Li, Ruohan Zhang, Josiah Wong, Cem Gokmen, Sanjana Srivastava, Roberto Martín-Martín, Chen Wang, Gabrael Levine, Wensi Ai, Benjamin Martinez, Hang Yin, Michael Lingelbach, Minjune Hwang, Ayano Hiranaka, Sujay Garlanka, Arman Aydin, Sharon Lee, Jiankai Sun, Mona Anvari, Manasi Sharma, Dhruva Bansal, Samuel Hunter, Kyu-Young Kim, Alan Lou, Caleb~R Matthews, Ivan Villa-Renteria, Jerry~Huayang Tang, Claire Tang, Fei Xia, Yunzhu Li, Silvio Savarese, Hyowon Gweon, C.~Karen Liu, Jiajun Wu, and Li Fei-Fei.
\newblock Behavior-1k: A human-centered, embodied ai benchmark with 1,000 everyday activities and realistic simulation.
\newblock \emph{arXiv preprint arXiv:2403.09227}, 2024.

\bibitem[Li et~al.(2025)Li, Zheng, He, Liu, Lin, Yang, Chen, and Guo]{li20252}
Haodong Li, Wangguangdong Zheng, Jing He, Yuhao Liu, Xin Lin, Xin Yang, Ying-Cong Chen, and Chunchao Guo.
\newblock Da: Depth anything in any direction.
\newblock \emph{arXiv preprint arXiv:2509.26618}, 2025.

\bibitem[Li et~al.(2022)Li, Jin, Hu, Dai, Du, and Li]{Li_Jin2022MODE}
Ming Li, Xueqian Jin, Xuejiao Hu, Jingzhao Dai, Sidan Du, and Yang Li.
\newblock Mode: Multi-view omnidirectional depth estimation with 360$^\circ$ cameras.
\newblock In \emph{European Conference on Computer Vision (ECCV)}, 2022.

\bibitem[Lin et~al.(2014)Lin, Maire, Belongie, Hays, Perona, Ramanan, Doll{\'a}r, and Zitnick]{lin2014microsoft}
Tsung-Yi Lin, Michael Maire, Serge Belongie, James Hays, Pietro Perona, Deva Ramanan, Piotr Doll{\'a}r, and C~Lawrence Zitnick.
\newblock Microsoft coco: Common objects in context.
\newblock In \emph{European conference on computer vision}, pages 740--755. Springer, 2014.

\bibitem[Lin et~al.(2025{\natexlab{a}})Lin, Ge, Zhang, Wan, Wang, Li, Jiang, Du, Tao, Yang, and Qi]{lin2025flightgapsurveyperspective}
Xin Lin, Xian Ge, Dizhe Zhang, Zhaoliang Wan, Xianshun Wang, Xiangtai Li, Wenjie Jiang, Bo Du, Dacheng Tao, Ming-Hsuan Yang, and Lu Qi.
\newblock One flight over the gap: A survey from perspective to panoramic vision, 2025{\natexlab{a}}.

\bibitem[Lin et~al.(2025{\natexlab{b}})Lin, Ge, Zhang, Wan, Wang, Li, Jiang, Du, Tao, Yang, et~al.]{lin2025one}
Xin Lin, Xian Ge, Dizhe Zhang, Zhaoliang Wan, Xianshun Wang, Xiangtai Li, Wenjie Jiang, Bo Du, Dacheng Tao, Ming-Hsuan Yang, et~al.
\newblock One flight over the gap: A survey from perspective to panoramic vision.
\newblock \emph{arXiv preprint arXiv:2509.04444}, 2025{\natexlab{b}}.

\bibitem[Liu et~al.(2023)Liu, Zhang, Qi, Wang, Zhang, and Wu]{liu2023aerialvln}
Shubo Liu, Hongsheng Zhang, Yuankai Qi, Peng Wang, Yanning Zhang, and Qi Wu.
\newblock Aerialvln: Vision-and-language navigation for uavs.
\newblock In \emph{Proceedings of the IEEE/CVF International Conference on Computer Vision}, pages 15384--15394, 2023.

\bibitem[Mellinger and Kumar(2011)]{mellinger2011minimum}
Daniel Mellinger and Vijay Kumar.
\newblock Minimum snap trajectory generation and control for quadrotors.
\newblock In \emph{2011 IEEE international conference on robotics and automation}, pages 2520--2525. IEEE, 2011.

\bibitem[Misra et~al.(2018)Misra, Bennett, Blukis, Niklasson, Shatkhin, and Artzi]{misra2018mapping}
Dipendra Misra, Andrew Bennett, Valts Blukis, Eyvind Niklasson, Max Shatkhin, and Yoav Artzi.
\newblock Mapping instructions to actions in 3d environments with visual goal prediction.
\newblock \emph{arXiv preprint arXiv:1809.00786}, 2018.

\bibitem[Piccinelli et~al.(2025)Piccinelli, Sakaridis, Segu, Yang, Li, Abbeloos, and Gool]{piccinelli2025unik3duniversalcameramonocular}
Luigi Piccinelli, Christos Sakaridis, Mattia Segu, Yung-Hsu Yang, Siyuan Li, Wim Abbeloos, and Luc~Van Gool.
\newblock Unik3d: Universal camera monocular 3d estimation, 2025.

\bibitem[Pintore et~al.(2021)Pintore, Almansa, Agus, and Gobbetti]{pintore2021deep3dlayout}
Giovanni Pintore, Eva Almansa, Marco Agus, and Enrico Gobbetti.
\newblock Deep3dlayout: 3d reconstruction of an indoor layout from a spherical panoramic image.
\newblock \emph{TOG}, 2021.

\bibitem[Pizzi et~al.(2022)Pizzi, Roy, Ravindra, Goyal, and Douze]{pizzi2022self}
Ed Pizzi, Sreya~Dutta Roy, Sugosh~Nagavara Ravindra, Priya Goyal, and Matthijs Douze.
\newblock A self-supervised descriptor for image copy detection.
\newblock In \emph{CVPR}, 2022.

\bibitem[Puig et~al.(2023)Puig, Undersander, Szot, Cote, Yang, Partsey, Desai, Clegg, Hlavac, Min, et~al.]{puig2023habitat}
Xavier Puig, Eric Undersander, Andrew Szot, Mikael~Dallaire Cote, Tsung-Yen Yang, Ruslan Partsey, Ruta Desai, Alexander~William Clegg, Michal Hlavac, So~Yeon Min, et~al.
\newblock Habitat 3.0: A co-habitat for humans, avatars and robots.
\newblock \emph{arXiv preprint arXiv:2310.13724}, 2023.

\bibitem[Qi et~al.(2019)Qi, Jiang, Liu, Shen, and Jia]{qi2019amodal}
Lu Qi, Li Jiang, Shu Liu, Xiaoyong Shen, and Jiaya Jia.
\newblock Amodal instance segmentation with kins dataset.
\newblock In \emph{Proceedings of the IEEE/CVF Conference on Computer Vision and Pattern Recognition}, pages 3014--3023, 2019.

\bibitem[Qi et~al.(2022{\natexlab{a}})Qi, Kuen, Guo, Shen, Gu, Jia, Lin, and Yang]{qi2022high}
Lu Qi, Jason Kuen, Weidong Guo, Tiancheng Shen, Jiuxiang Gu, Jiaya Jia, Zhe Lin, and Ming-Hsuan Yang.
\newblock High-quality entity segmentation.
\newblock \emph{arXiv preprint arXiv:2211.05776}, 2022{\natexlab{a}}.

\bibitem[Qi et~al.(2022{\natexlab{b}})Qi, Kuen, Wang, Gu, Zhao, Torr, Lin, and Jia]{qi2022open}
Lu Qi, Jason Kuen, Yi Wang, Jiuxiang Gu, Hengshuang Zhao, Philip Torr, Zhe Lin, and Jiaya Jia.
\newblock Open world entity segmentation.
\newblock \emph{IEEE Transactions on Pattern Analysis and Machine Intelligence}, 45\penalty0 (7):\penalty0 8743--8756, 2022{\natexlab{b}}.

\bibitem[Qiu et~al.(2017)Qiu, Zhong, Zhang, Qiao, Xiao, Kim, and Wang]{qiu2017unrealcv}
Weichao Qiu, Fangwei Zhong, Yi Zhang, Siyuan Qiao, Zihao Xiao, Tae~Soo Kim, and Yizhou Wang.
\newblock Unrealcv: Virtual worlds for computer vision.
\newblock In \emph{ACM MM}, 2017.

\bibitem[Sekkat et~al.(2020)Sekkat, Dupuis, Vasseur, and Honeine]{sekkat2020omniscape}
Ahmed~Rida Sekkat, Yohan Dupuis, Pascal Vasseur, and Paul Honeine.
\newblock The omniscape dataset.
\newblock In \emph{2020 IEEE International conference on robotics and automation (ICRA)}, pages 1603--1608. IEEE, 2020.

\bibitem[Shah et~al.(2017{\natexlab{a}})Shah, Dey, Lovett, and Kapoor]{airsim2017fsr}
Shital Shah, Debadeepta Dey, Chris Lovett, and Ashish Kapoor.
\newblock Airsim: High-fidelity visual and physical simulation for autonomous vehicles.
\newblock In \emph{Field and Service Robotics}, 2017{\natexlab{a}}.

\bibitem[Shah et~al.(2017{\natexlab{b}})Shah, Dey, Lovett, and Kapoor]{shah2017airsim}
Shital Shah, Debadeepta Dey, Chris Lovett, and Ashish Kapoor.
\newblock Airsim: High-fidelity visual and physical simulation for autonomous vehicles.
\newblock In \emph{Field and service robotics: Results of the 11th international conference}, 2017{\natexlab{b}}.

\bibitem[Shi et~al.(2017)Shi, Li, Wang, Luo, Huang, and Fukuda]{shi2017design}
Qing Shi, Chang Li, Chunbao Wang, Haibo Luo, Qiang Huang, and Toshio Fukuda.
\newblock Design and implementation of an omnidirectional vision system for robot perception.
\newblock \emph{Mechatronics}, 41:\penalty0 58--66, 2017.

\bibitem[Sobchyshak et~al.(2025)Sobchyshak, Berrezueta-Guzman, and Wagner]{sobchyshak2025pushing}
Oleksandra Sobchyshak, Santiago Berrezueta-Guzman, and Stefan Wagner.
\newblock Pushing the boundaries of immersion and storytelling: A technical review of unreal engine.
\newblock \emph{Displays}, page 103268, 2025.

\bibitem[Wang et~al.(2024{\natexlab{a}})Wang, Yang, Li, Gou, Yan, Zeng, Gao, Wang, Jing, and Zhang]{wang2024freeman}
Jiong Wang, Fengyu Yang, Bingliang Li, Wenbo Gou, Danqi Yan, Ailing Zeng, Yijun Gao, Junle Wang, Yanqing Jing, and Ruimao Zhang.
\newblock Freeman: Towards benchmarking 3d human pose estimation under real-world conditions.
\newblock In \emph{Proceedings of the IEEE/CVF Conference on Computer Vision and Pattern Recognition}, pages 21978--21988, 2024{\natexlab{a}}.

\bibitem[Wang et~al.(2025)Wang, Li, Zhang, Yu, Yuan, She, Guo, Zheng, Howe, Chandra, et~al.]{wang2025uavscenes}
Sijie Wang, Siqi Li, Yawei Zhang, Shangshu Yu, Shenghai Yuan, Rui She, Quanjiang Guo, JinXuan Zheng, Ong~Kang Howe, Leonrich Chandra, et~al.
\newblock Uavscenes: A multi-modal dataset for uavs.
\newblock In \emph{ICCV}, 2025.

\bibitem[Wang et~al.(2024{\natexlab{b}})Wang, Yang, Wang, Kwan, Chen, Wu, Li, Liao, and Liu]{wang2024towards}
Xiangyu Wang, Donglin Yang, Ziqin Wang, Hohin Kwan, Jinyu Chen, Wenjun Wu, Hongsheng Li, Yue Liao, and Si Liu.
\newblock Towards realistic uav vision-language navigation: Platform, benchmark, and methodology.
\newblock \emph{arXiv preprint arXiv:2410.07087}, 2024{\natexlab{b}}.

\bibitem[Xu et~al.(2022)Xu, Zhao, Ma, Li, Yuan, Feng, Yan, and Dai]{xu2022pandora}
Hang Xu, Qiang Zhao, Yike Ma, Xiaodong Li, Peng Yuan, Bailan Feng, Chenggang Yan, and Feng Dai.
\newblock Pandora: A panoramic detection dataset for object with orientation.
\newblock In \emph{European conference on computer vision}, pages 237--252. Springer, 2022.

\bibitem[Xu et~al.(2018)]{xu2018predicting}
Menghan Xu et~al.
\newblock Predicting head movement in panoramic video: A deep reinforcement learning approach.
\newblock In \emph{CVPR}, 2018.

\bibitem[Yang et~al.(2019)Yang, Hu, Bergasa, Romera, and Wang]{yang2019pass}
Kailun Yang, Xinxin Hu, Luis~M Bergasa, Eduardo Romera, and Kaiwei Wang.
\newblock Pass: Panoramic annular semantic segmentation.
\newblock \emph{IEEE Transactions on Intelligent Transportation Systems}, 21\penalty0 (10):\penalty0 4171--4185, 2019.

\bibitem[Yang et~al.(2021)Yang, Zhang, Rei{\ss}, Hu, and Stiefelhagen]{yang2021capturing}
Kailun Yang, Jiaming Zhang, Simon Rei{\ss}, Xinxin Hu, and Rainer Stiefelhagen.
\newblock Capturing omni-range context for omnidirectional segmentation.
\newblock In \emph{CVPR}, 2021.

\bibitem[Zavlangas et~al.(2000)Zavlangas, Tzafestas, and Althoefer]{zavlangas2000fuzzy}
Panagiotis~G Zavlangas, Spyros~G Tzafestas, and Kasper Althoefer.
\newblock Fuzzy obstacle avoidance and navigation for omnidirectional mobile robots.
\newblock In \emph{European Symposium on Intelligent Techniques, Aachen, Germany}, pages 375--382, 2000.

\bibitem[Zheng et~al.(2024)Zheng, Liu, Chen, Peng, Wu, Yang, Zhang, and Stiefelhagen]{zheng2024open}
Junwei Zheng, Ruiping Liu, Yufan Chen, Kunyu Peng, Chengzhi Wu, Kailun Yang, Jiaming Zhang, and Rainer Stiefelhagen.
\newblock Open panoramic segmentation.
\newblock In \emph{ECCV}, 2024.

\bibitem[Zhong et~al.(2025)Zhong, Wu, Wang, Chen, Ci, Li, and Wang]{zhong2025unrealzoo}
Fangwei Zhong, Kui Wu, Churan Wang, Hao Chen, Hai Ci, Zhoujun Li, and Yizhou Wang.
\newblock Unrealzoo: Enriching photo-realistic virtual worlds for embodied ai.
\newblock In \emph{ICCV}, 2025.

\bibitem[Zhou et~al.(2019)Zhou, Zhao, Puig, Xiao, Fidler, Barriuso, and Torralba]{zhou2019semantic}
Bolei Zhou, Hang Zhao, Xavier Puig, Tete Xiao, Sanja Fidler, Adela Barriuso, and Antonio Torralba.
\newblock Semantic understanding of scenes through the ade20k dataset.
\newblock \emph{International Journal of Computer Vision}, 127\penalty0 (3):\penalty0 302--321, 2019.

\end{thebibliography}
}

\clearpage
\appendix
\section*{Appendix}
This Supplementary Material provides technical details, comprehensive dataset statistics, and additional implementation specifics that were omitted from the main paper due to space constraints. Specifically, we present the following information:
\begin{itemize}
    \item In Section \ref{sec:dataset_stats}, we provide some details of the three subsets of the Omni360-X dataset: Omni360-Scene, Omni360-Human, and Omni360-WayPoint, including the semantic categories and pedestrian behaviors.
    \item In Section \ref{sec:min_snap}, we elaborate on the mathematical model and constraints of the Minimum Snap trajectory planning method used for automated trajectory generation.
    \item In Section \ref{sec:exp_details}, we provide comprehensive experimental configurations for the Monocular Pedestrian Distance Estimation (MPDE) and Panoramic Vision-Language Navigation (VLN) tasks.
\end{itemize}

Also, please check \underline{\textit{a recorded video}} to obtain a brief description of our paper.

\section{More Details of Omni360-X Dataset}
\label{sec:dataset_stats}


\subsection{Omni360-Scene Statistics}
Omni360-Scene provides pixel-level annotations for depth information, semantic segmentation, and entity segmentation across diverse environments. As semantic complexity varies by scene, Table \ref{tab:semantic_categories} provides a visual overview including the ERP image and semantic segmentation mask for each scenario, followed by a detailed breakdown of the semantic categories.

\begin{table*}[t]
\centering
\caption{Visualization of semantic segmentation and list of semantic categories for each scene.}
\label{tab:semantic_categories}
\small
\renewcommand{\arraystretch}{1.2} 
\resizebox{\textwidth}{!}{%
\begin{tabular}{>{\centering\arraybackslash}m{2.5cm} 
                >{\centering\arraybackslash}m{3cm} 
                >{\centering\arraybackslash}m{3cm} 
                >{\centering\arraybackslash}m{6.5cm}}
\toprule
\textbf{Scene Name} & \textbf{ERP Image} & \textbf{Semantic Vis} & \textbf{Semantic Categories} \\
\midrule
\textbf{City Park} & 
\includegraphics[width=\linewidth,valign=c]{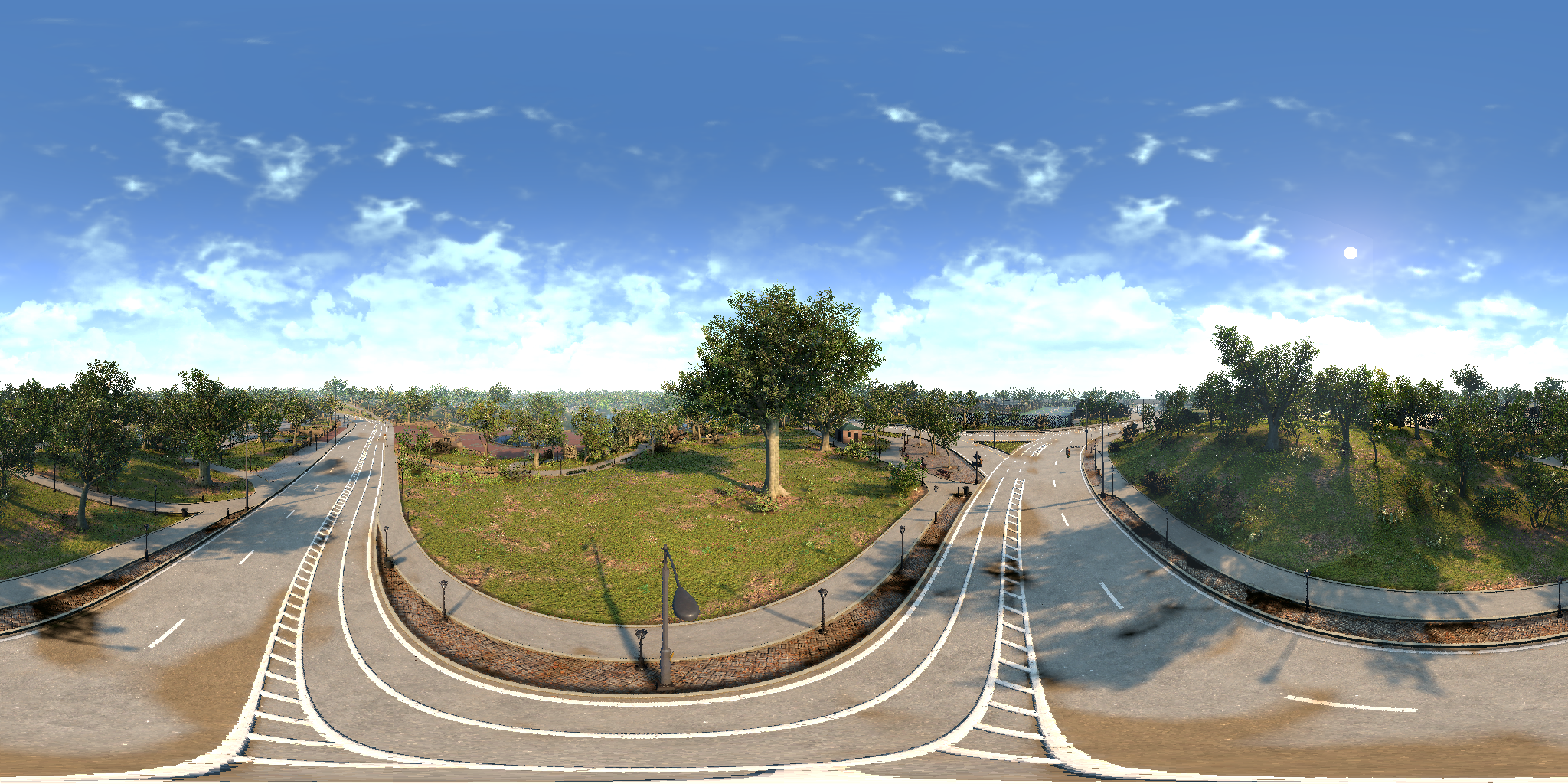} & 
\includegraphics[width=\linewidth,valign=c]{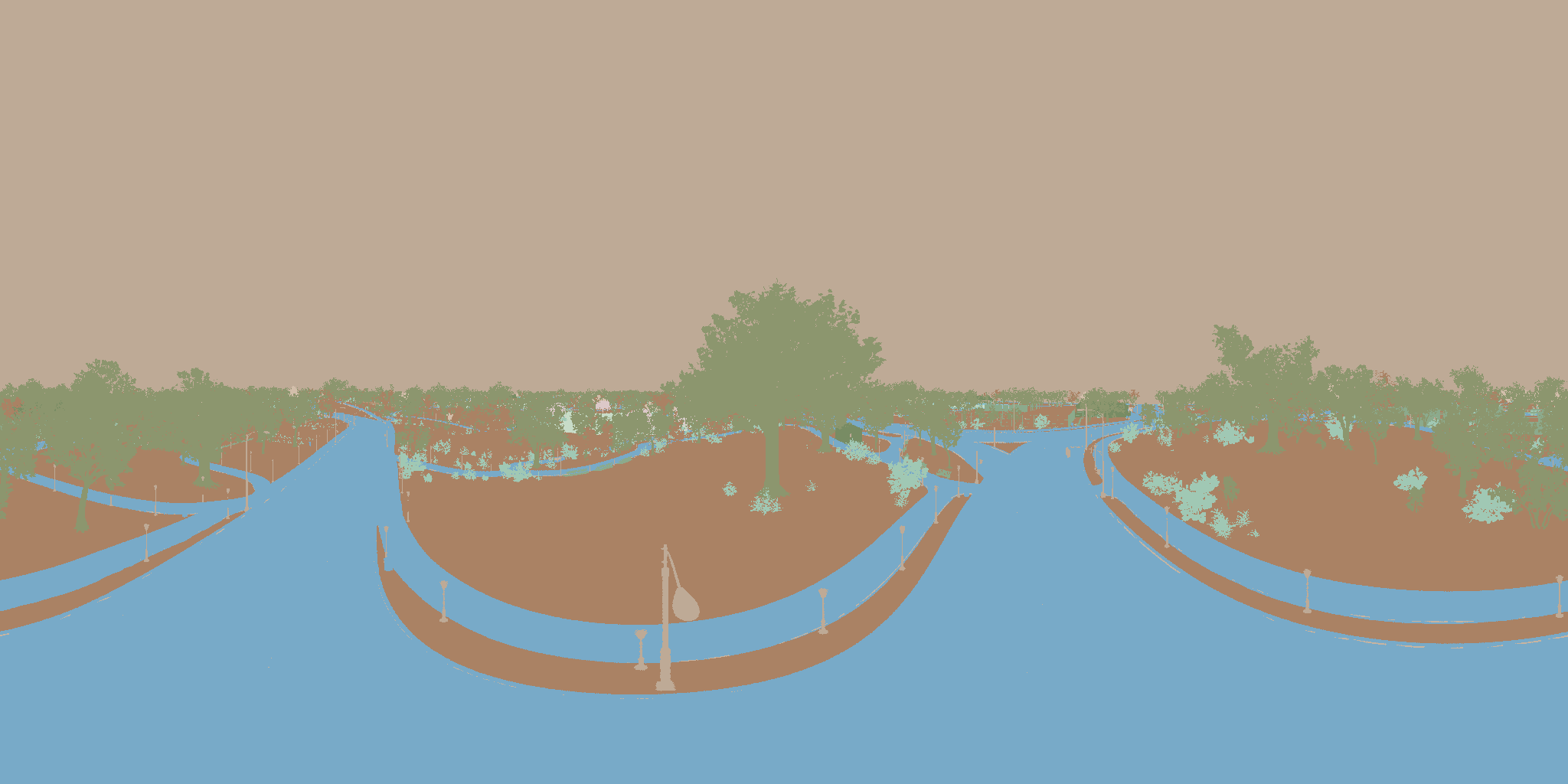} & 
Building, Rock, AmurCork, Bush, Elm, Ivy, Maple, WeepingWillow, PlayGround, Bench, LampPost, FoodStalls, Cafechair, Roadblock, Trashcan, Trafficbarrel, Circlefence, Trafficlight, Water Plane, Road, Cafetable, Umbrella, Pool Sidewalk, Sky, Landscape \\
\midrule
\textbf{Downtown West} & 
\includegraphics[width=\linewidth,valign=c]{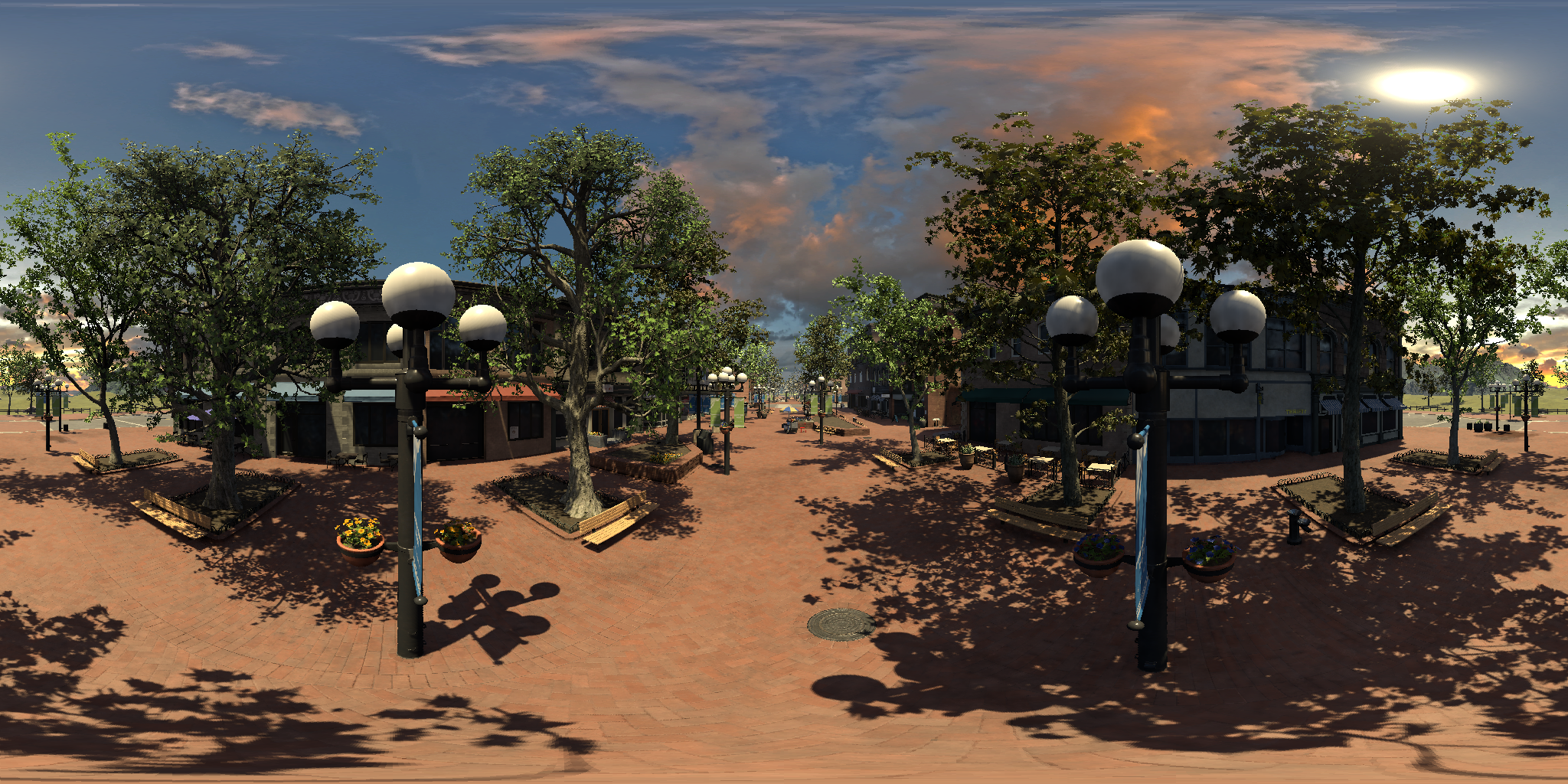} & 
\includegraphics[width=\linewidth,valign=c]{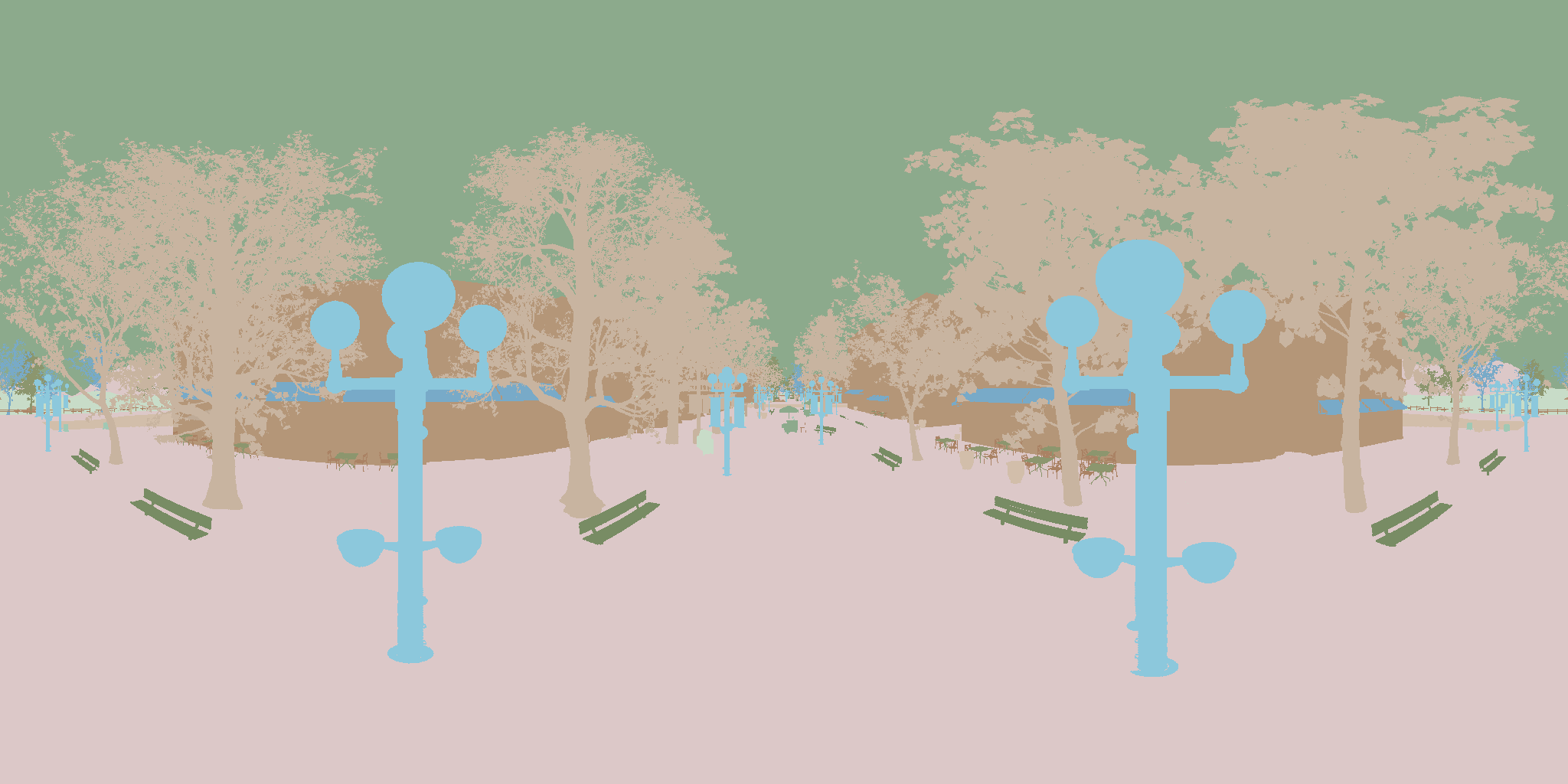} & 
Building, Awning, Roof, Tree Generic, Tree Narrowleaf, Tree Pine, Prop Dining Table, Umbrella, Prop Dining Chair, Pot, Car Pillar, Recycle Bin, Food Cart, Bench Wood, Poster Stand, Ground Mod, Road, Lightpost Light Post, Tarppost, Light Streetlight Complete, Tarponly, Ground Park Walkway, Rock Rock, Background Mountains, Wood Fence Wood Fence, Prop Park Railing Rail, Prop Park Railing Pillar, Sky, Landscape \\
\midrule
\textbf{SF City} & 
\includegraphics[width=\linewidth,valign=c]{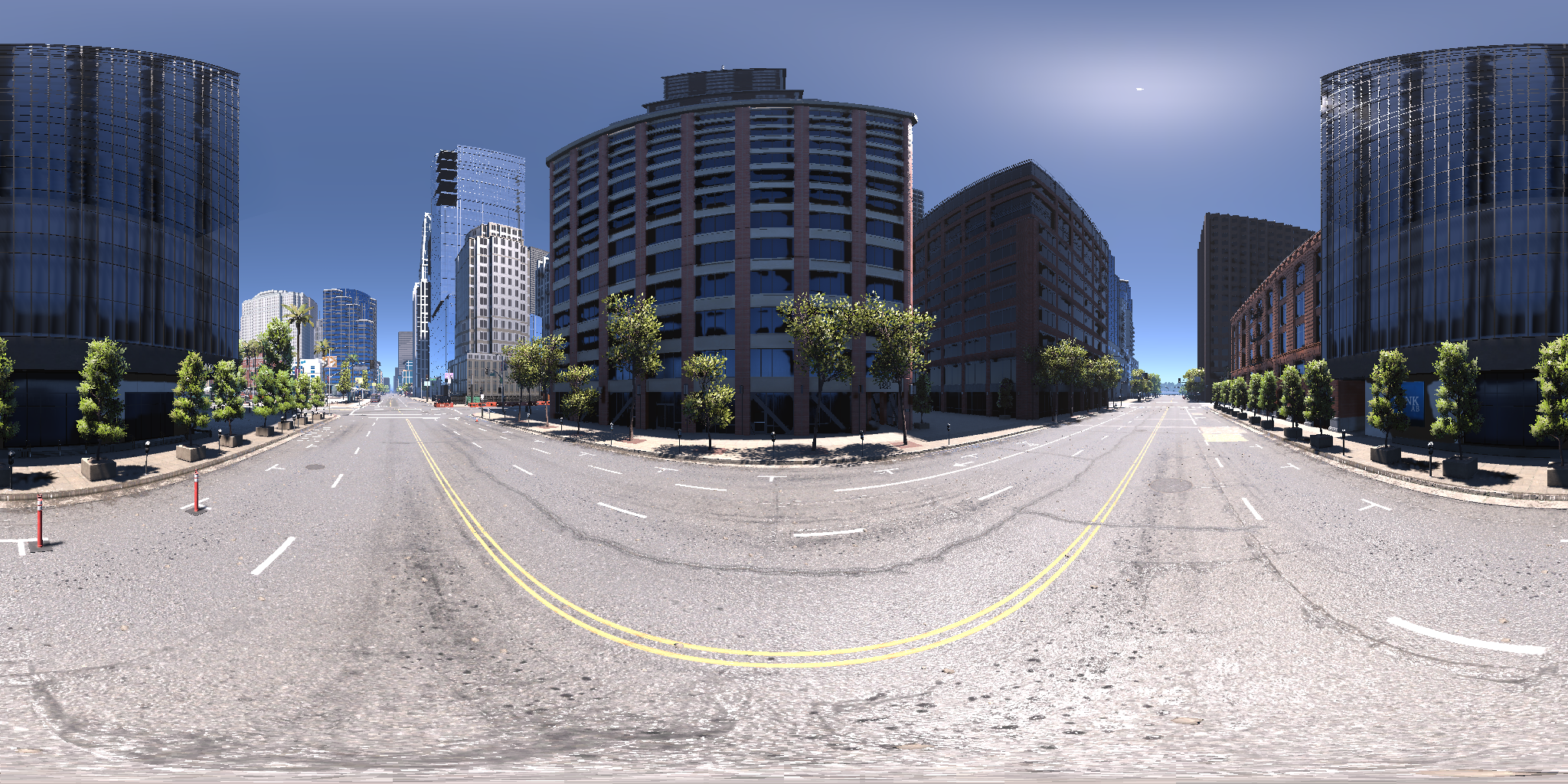} & 
\includegraphics[width=\linewidth,valign=c]{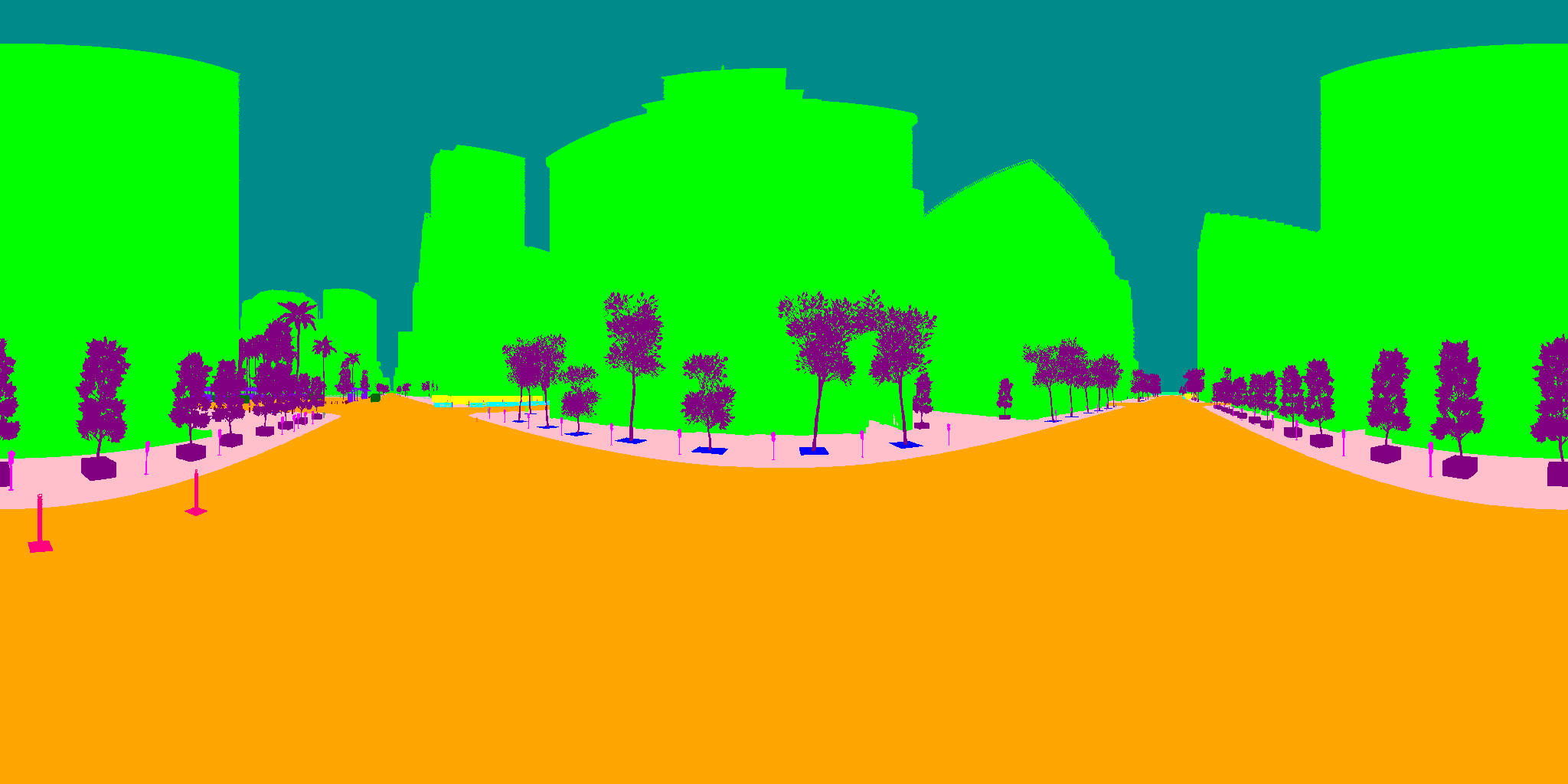} & 
Building, Sidewalk, Road, Bus, Fence, Cone, Hydrant, Parkingmeter, Stopstation, Elecbox, Trash, Traffictube, Barrier, Alamppost, Blamppost, Lake, Bollardrope, Barricademetal, Tree, Sky \\
\midrule
\textbf{New York City} & 
\includegraphics[width=\linewidth,valign=c]{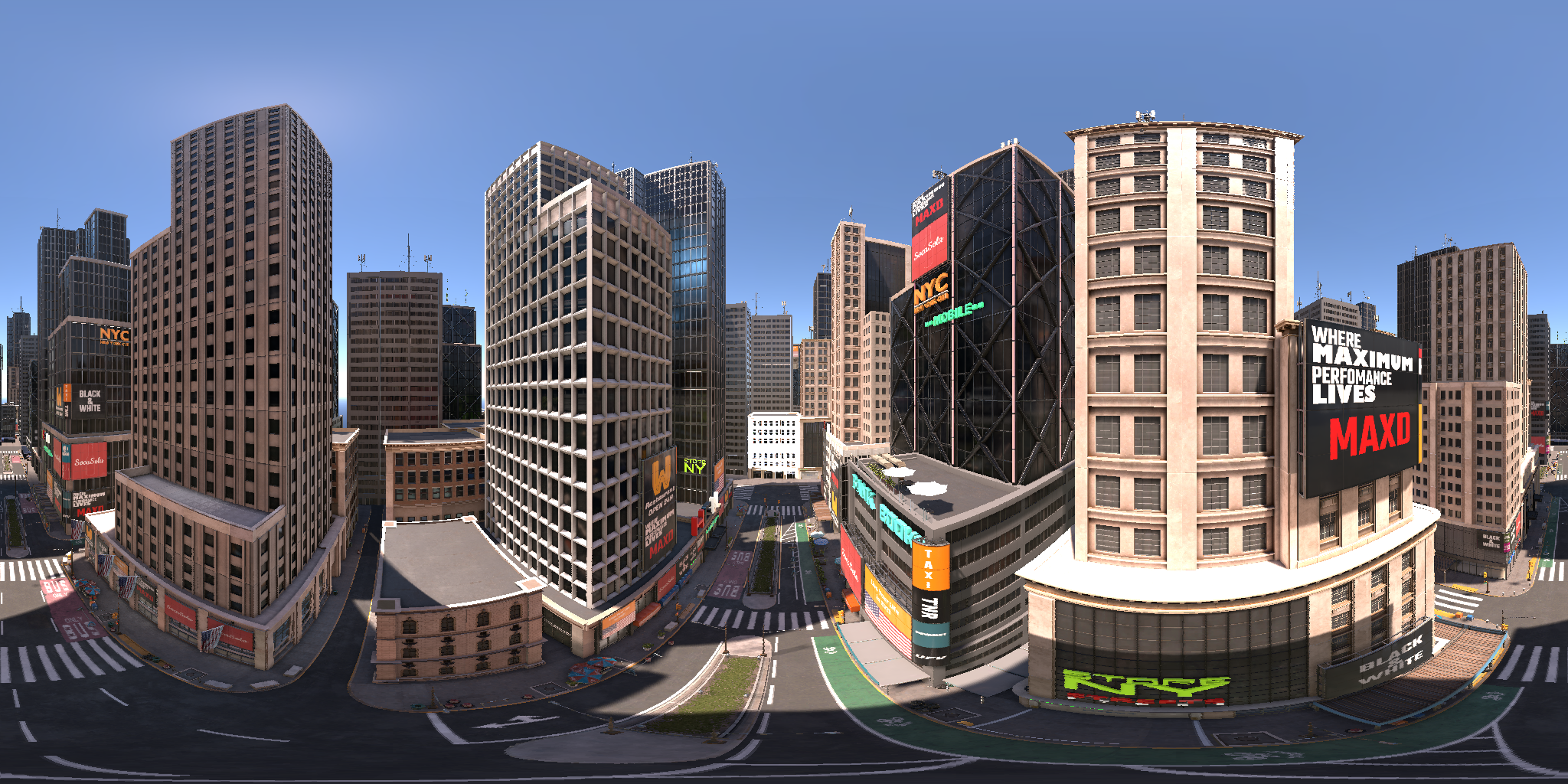} & 
\includegraphics[width=\linewidth,valign=c]{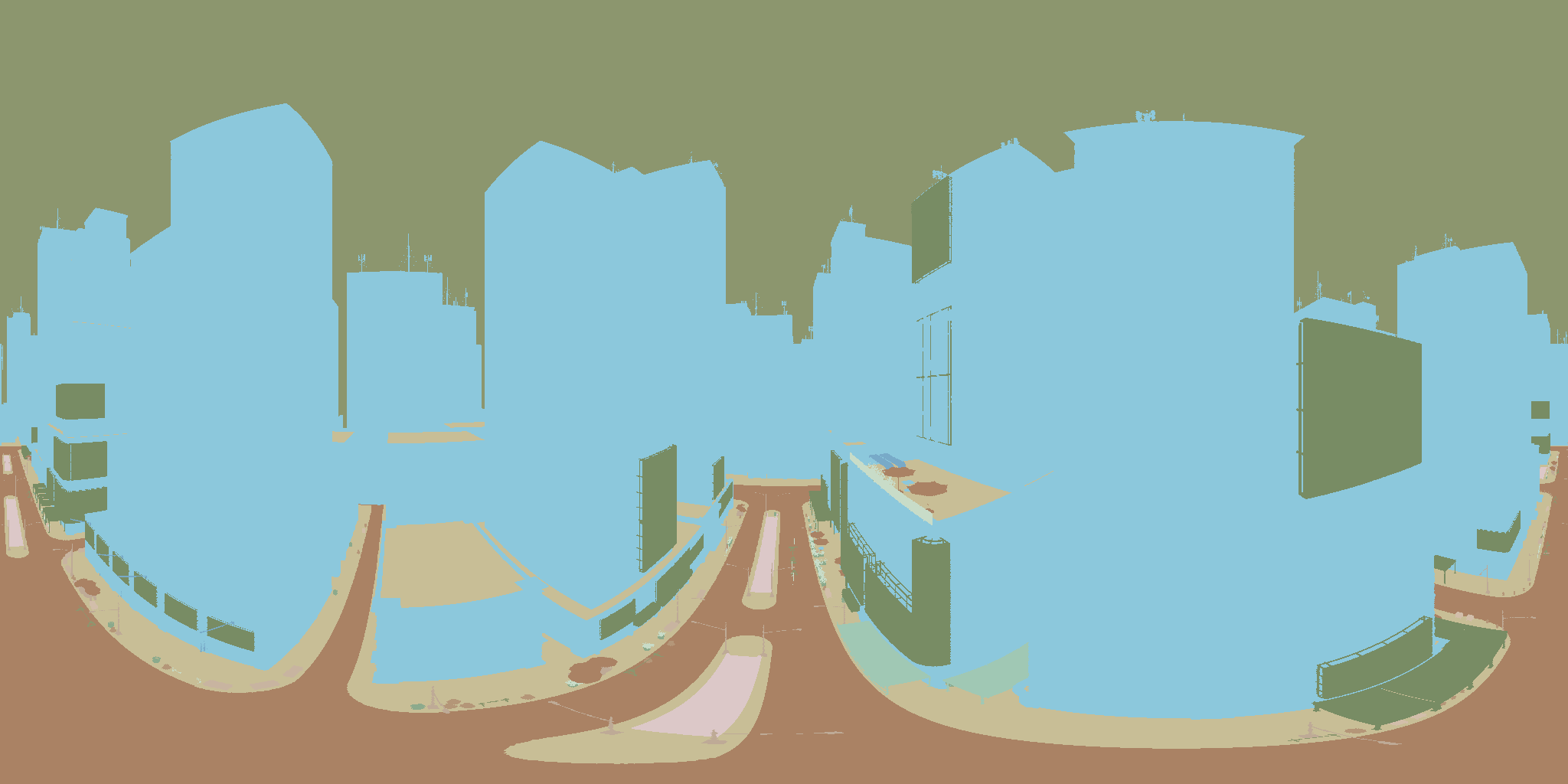} & 
Concreteblock, Streetprops, Plasticcone, Metalfence, Pillar, Lamp, Trashcan, Postbox, Umbrella, Table, Chair, Greenpot, Roadcolumn, Adplane, Buidlingawning, Scaffolding, Usaflag, Plant, Ventilationtube, Building, Hotdogpot, Road, Sidewalk, Grounddirt, Sky \\
\bottomrule
\end{tabular}%
}
\end{table*}

\subsection{Omni360-Human Statistics}
The Omni360-Human subset is dedicated to human-centric perception tasks, primarily monocular pedestrian distance estimation. 

\textbf{Data Statistics:} Table \ref{tab:human_stats} presents a detailed breakdown of the dataset composition. The data was collected across 6 distinct scenarios , covering a wide range of crowd densities and area sizes. The dataset includes over 100K frames in total, with varying numbers of NPCs to simulate realistic crowd dynamics.

\begin{table}[!ht]
\centering
\caption{Detailed Statistics of the Omni360-Human Dataset.}
\label{tab:human_stats}
\resizebox{\columnwidth}{!}{
\begin{tabular}{l c c c c}
\toprule
\textbf{Scene} & \textbf{Subsets} & \textbf{\makecell{Area Range \\($m \times m$)}} & \textbf{\makecell{NPC Count \\(Min-Max)}} & \textbf{\makecell{Total Frames}} \\
\midrule
New York City & 14 & $12\times12 \sim 30\times30$ & $15 \sim 45$ & 29,000 \\
LisbonDowtown & 10 & $12\times12 \sim 30\times50$ & $10 \sim 45$ & 9,000 \\
Downtown City & 17 & $12\times12 \sim 30\times30$ & $8 \sim 30$ & 27,000 \\
Roof & 7 & $12\times12 \sim 45\times20$ & $5 \sim 30$ & 11,200 \\
Rural Cabins & 2 & $15\times15 \sim 15\times30$ & $7 \sim 14$ & 4,000 \\
Rome & 11 & $8\times10 \sim 50\times30$ & $4 \sim 30$ & 20,500 \\
\midrule
\textbf{Total} & \textbf{61} & \textbf{-} & \textbf{-} & \textbf{100,700} \\
\bottomrule
\end{tabular}}
\end{table}

\subsection{Omni360-WayPoint Statistics}
Omni360-WayPoint provides physics-consistent UAV flight paths for navigation, trajectory prediction, and control. The trajectories adhere to realistic flight dynamics derived from Minimum Snap planning. Table \ref{tab:min_snap_params} and Table \ref{tab:intro_wapts} details the key kinematic parameters and scale of the waypoint data.

\begin{table}[!ht]
\centering
\caption{Introduction of Omni360-WayPoint. The Kinematic Parameters include two distinct sets of {$a_{\max}$, $v_{\max}$, sampling interval $t$}, each representing a typical UAV flight condition. The Total Number of Flight Paths is computed as the product of the number of Kinematic Parameter sets and the Number of Routes.}
\label{tab:intro_wapts}
\resizebox{\columnwidth}{!}{
\begin{tabular}{lcccc}
\hline
\textbf{Scenario} &
\makecell{\textbf{Length}\\\textbf{Range}} &
\makecell{\textbf{Kinematic}\\\textbf{Parameters}} &
\makecell{\textbf{Number}\\\textbf{of Routes}} &
\makecell{\textbf{Total Number}\\\textbf{of Flight Paths}} \\ \hline
City Park & [50, 150] & [(3, 16, 0.5), (5, 21, 1)] & 20000 & 40000 \\
Downtown West & [20, 50] & [(3, 16, 0.5), (5, 21, 1)] & 5000 & 10000 \\
New York City & [20, 50] & [(3, 16, 0.5), (5, 21, 1)] & 5000 & 10000 \\
SF City & [50, 150] & [(3, 16, 0.5), (5, 21, 1)] & 20000 & 40000 \\ \hline
\end{tabular}}
\end{table}

\begin{table}[htbp]
    \caption{Inputs and outputs of the trajectory way-points generation algorithm. 
The input $v_{\max}$ denotes the predefined maximum flight speed, $a_{\max}$ represents the maximum aircraft acceleration, and $t$ is the sampling interval. The output parameters are also described in Eq.~(\ref{eq:minimum}).}
    \label{tab:min_snap_params}
    \centering
    \begin{tabular}{lccc}
        \toprule
        Minimum Snap & \multicolumn{3}{c}{Parameters} \\
        \cmidrule(lr){2-4}
        Input & $v_{\max}$ & $a_{\max}$ & $t$ \\
        \midrule
        Output & $\mathbf{p}(t)$ & $\mathbf{v}(t)$ & $\mathbf{a}(t)$ \\
        \bottomrule
    \end{tabular}
\end{table}

\begin{equation}
    \label{eq:minimum}
    \mathbf{S}(t) = 
    \begin{bmatrix}
        \mathbf{p}(t)^T \\
        \mathbf{v}(t)^T \\
        \mathbf{a}(t)^T
    \end{bmatrix}
    =
    \begin{bmatrix}
        x(t) & y(t) & z(t) \\
        v_x(t) & v_y(t) & v_z(t) \\
        a_x(t) & a_y(t) & a_z(t)
    \end{bmatrix}
\end{equation}


\section{Minimum Snap Trajectory Planning Implementation Details}
\label{sec:min_snap}
The Automated Trajectory Generation Paradigm employs Minimum Snap trajectory planning to produce smooth, dynamically feasible UAV flight paths from sparse user-defined waypoints. This method minimizes the integrated square of the fourth derivative of position (Snap), effectively ensuring trajectory smoothness and reduced control effort.

\paragraph{Polynomial Representation.}
Given a sequence of key waypoints $\{p_0, p_1, \dots, p_M\}$, each segment of the trajectory is modeled as a fifth-order polynomial:
\begin{equation}
p_i(t) = a_{i,0} + a_{i,1}t + a_{i,2}t^2 + a_{i,3}t^3 + a_{i,4}t^4 + a_{i,5}t^5,
\label{eq:poly}
\end{equation}
where $\mathbf{a}_i = [a_{i,0}, \dots, a_{i,5}]^\top$ are the polynomial coefficients for segment $i$.

\paragraph{Optimization Objective.}
Following the Minimum Snap formulation, the smoothness of the trajectory is achieved by minimizing the integral of the squared fourth derivative (snap):
\begin{equation}
J = \int_{t_0}^{t_M} \left\| \frac{d^4 p(t)}{dt^4} \right\|^2 dt.
\label{eq:snap}
\end{equation}

\begin{table*}[!htp]
\centering
\small
\caption{MPDE results across different training and testing datasets. \textbf{Dist. Err (All)} denotes the weighted average over all four datasets, where \textbf{Dist} refers to the Euclidean distance. \textbf{Ang. Err (All)} denotes the weighted average over all four datasets, where \textbf{Ang} refers to the angle. The \textbf{Pub} column in the last two columns indicates that the weighted average is computed over only the top three public datasets.
}
\label{tab:cross_dataset}
\renewcommand{\arraystretch}{1.15}
\setlength{\tabcolsep}{3.5pt}

\begin{tabular}{
    >{\centering\arraybackslash}m{2.7cm}  
    >{\centering\arraybackslash}m{2.5cm}  
    >{\centering\arraybackslash}m{1.3cm}  
    >{\centering\arraybackslash}m{1.3cm}  
    >{\centering\arraybackslash}m{1.2cm}  
    >{\centering\arraybackslash}m{1.1cm}  
    >{\centering\arraybackslash}m{1.4cm}  
    >{\centering\arraybackslash}m{1.4cm}  
    >{\centering\arraybackslash}m{1.4cm}  
}
\toprule
\textbf{\makecell{Training\\Set}} &
\textbf{\makecell{Test\\Set}} &
\textbf{\makecell{Dist.\\Err}} &
\textbf{\makecell{Samples}} &
\textbf{\makecell{Dist. Err\\(All)}} &
\textbf{\makecell{Ang.\\Err}} &
\textbf{\makecell{Ang. Err\\(All)}} &
\textbf{\makecell{Ang. Err\\(Pub)}} &
\textbf{\makecell{Dist. Err\\(Pub)}} \\
\midrule

\multirow{4}{*}{\centering\makecell{nuScenes}} 
 & nuScenes & 1.078 & 15369 & \multirow{4}{*}{\centering 0.80} & 31.90 & \multirow{4}{*}{\centering 23.14} & \multirow{4}{*}{\centering 21.207} & \multirow{4}{*}{\centering 0.484} \\
 & KITTI & 0.822 & 1759 &  & 31.50 &  &  &  \\
 & FreeMan & 0.260 & 43361 &  & 17.00 &  &  &  \\
 & Omni360-Human & 2.439 & 11496 &  & 33.30 &  &  &  \\

\midrule

\multirow{4}{*}{\centering\makecell{nuScenes\\+ Omni360-\\Human-all}} 
 & nuScenes & 1.073 & 15369 & \multirow{4}{*}{\centering \textcolor{red}{\textbf{0.43}}} & 30.70 & \multirow{4}{*}{\centering \textcolor{red}{\textbf{16.25}}} & \multirow{4}{*}{\centering 17.282} & \multirow{4}{*}{\centering 0.449} \\
 & KITTI & 0.802 & 1759  &  & 32.70 &  &  &  \\
 & FreeMan & 0.213 & 43361 &  & 11.90 &  &  &  \\
 & Omni360-Human & 0.313 & 11496 &  & 10.80 &  &  &  \\

\midrule

\multirow{4}{*}{\centering\makecell{nuScenes\\+ Omni360-\\Human-pitch\_0}} 
 & nuScenes & 1.071 & 15369 & \multirow{4}{*}{\centering 0.50} & 30.70 & \multirow{4}{*}{\centering 19.08} & \multirow{4}{*}{\centering 19.194} & \multirow{4}{*}{\centering \textcolor{red}{\textbf{0.433}}} \\
 & KITTI & 0.812 & 1759  &  & 31.90 &  &  &  \\
 & FreeMan & 0.191 & 43361 &  & 14.60 &  &  &  \\
 & Omni360-Human & 0.868 & 11496 &  & 18.50 &  &  &  \\

\midrule

\multirow{4}{*}{\centering\makecell{nuScenes\\+ Omni360-\\Human-pitch\_20}} 
 & nuScenes & 1.068 & 15369 & \multirow{4}{*}{\centering 0.67} & 30.70 & \multirow{4}{*}{\centering 16.73} & \multirow{4}{*}{\centering \textcolor{red}{\textbf{17.023}}} & \multirow{4}{*}{\centering 0.458} \\
 & KITTI & 0.809 & 1759  &  & 31.20 &  &  &  \\
 & FreeMan & 0.228 & 43361 &  & 11.60 &  &  &  \\
 & Omni360-Human & 1.779 & 11496 &  & 15.20 &  &  &  \\

\bottomrule
\end{tabular}
\end{table*}


\paragraph{Quadratic Programming Formulation.}
The optimization problem can be expressed as a quadratic program:
\begin{equation}
\begin{aligned}
\min_{\mathbf{a}} \quad & \mathbf{a}^\top Q \mathbf{a}, \\
\text{s.t.} \quad & A\mathbf{a} = b,
\end{aligned}
\label{eq:qp}
\end{equation}
where $Q$ is derived from the cost in~\eqref{eq:snap}, and $A$, $b$ encode waypoint and continuity constraints up to the third derivative. 
Solving this system yields the polynomial coefficients $\mathbf{a}$ defining the minimum-snap trajectory.

\paragraph{Dynamic Feasibility.}
To ensure physical feasibility, the trajectory is further constrained by dynamic limits on velocity and acceleration:
\begin{equation}
\|\dot{p}(t)\| \le v_{\max}, \qquad \|\ddot{p}(t)\| \le a_{\max}.
\label{eq:limits}
\end{equation}
Each segment duration $\Delta T_i$ is automatically adjusted according to these limits, as well as the chosen sampling interval $\Delta t$, ensuring that the resulting trajectory remains dynamically executable by the UAV controller.

\section{Experimental Details}
\label{sec:exp_details}

This section provides additional implementation details and experimental settings for the Monocular Pedestrian Distance Estimation (MPDE) and Panoramic Vision-Language Navigation (VLN) tasks presented in the main paper.

\subsection{Monocular Pedestrian Distance Estimation (MPDE)}

In the Monocular Pedestrian Distance Estimation experiments, we design four sets of evaluations to demonstrate the effectiveness of our data. We first report the results on all test sets using only the nuScenes dataset. We then conduct a series of comparative experiments on three configurations of the Omni360-Human dataset: the full dataset, the subset with a pitch angle of 0°, and the subset with a pitch angle of 20°.

All models are trained using the AdamW optimizer with an initial learning rate of 0.002 and a weight decay coefficient of 0.01. The learning rate is multiplied by 0.98 every 300 steps during training.

The Omni360-Human training set is curated to exclude any samples from the Omni360-Human test set used in the experiments.

\begin{table*}[t]
\centering
\caption{List of Prompts. The following prompts are used in two different environments, namely the \href{https://www.fab.com/listings/fb179131-13b6-47e3-823e-b8d9f48f3391}{New York City} scene and the \href{https://www.fab.com/listings/221af99d-8805-44b7-8844-933950f1ceb8}{1950s NYC Environment Megapack scene}.}
\label{tab:prompts}
\resizebox{\textwidth}{!}{
\begin{tabular}{>{\centering\arraybackslash}p{\textwidth}}
\hline
\textbf{Prompt} \\ \hline
Find the nearest traffic light and stop when you reach it. \\
Find the nearest blue mailbox and stop when you reach it. \\
Find the nearest tall building straight ahead and stop when you reach its rooftop. \\
Move forward, then at the intersection, you'll see a red telephone booth on your right. Stop near the closest one. \\
Fly across the lake in front of you, reach the opposite neighborhood, and stop on the street. \\
Find the nearby lake surrounded by woods and stop at the nearest shore. \\
Locate the building nearby with a giant Coca-Cola bottle decoration on its roof and fly close to the decoration. \\
Cross the zebra crossing and fly over this section of the road. \\
Fly to the small island in the center of the lake and land. \\
Fly straight ahead, turn right at the intersection, and stop near the bridge. \\
Fly along the current road and stop when you reach the second tree. \\
Stop near the small fountain located downstairs in the nearby building. \\
You are currently on the left side of the bridge. Now move to the right side and stop. \\
Climb over the fence in front of you and stop on the path in the park ahead. \\
Fly to the blue billboard on the building ahead and to your left, then stop nearby. \\
Fly to the tree with red leaves on the left side of the street and stop nearby. \\
Fly to the vicinity of the three very similar buildings straight ahead and stop. \\
Locate the billboard straight ahead featuring a person in a blue suit, fly to it, and stop nearby. \\
Fly to the billboard with the red car on the building ahead above you and stop nearby. \\
Find the floor in the building in front of you with red curtains and stop nearby. \\
Locate the billboard with the black and white portrait on the building ahead and stop nearby. \\
Arrive at the bank with the purple sign and stop downstairs. \\
Find the nearest yellow sunshade among the many downstairs and stop there. \\
Fly along the crosswalk over the intersection and stop on the opposite side of the road. \\
Find the nearest American flag and stop there. \\
Continue flying straight ahead and stop when you reach the intersection with the main road. \\
Fly to the blue barrier ahead and stop. \\
Fly to the red phone booth behind of you and stop when you are close the red phone booth. \\
Fly to the blue billboard on your right rear and stop when you’re close to it. \\
Fly to the lake surrounded by trees on your right and stop when you’re close to it. \\
Fly to the red bridge on your right and stop when you’re close to it. \\
Stop near the nearest lawn. \\
Find the nearest traffic light and stop near it. \\
Navigate to the nearest green bike lane and stop. \\
Fly to the nearest billboard that shows BLACK \& WHITE and stop nearby. \\
Navigate to the orange mailbox ahead and stop nearby. \\
Fly to the nearest food truck and stop. \\
\hline
\end{tabular}
}
\end{table*}

\subsection{Panoramic Vision-Language Navigation (VLN)}

In Visual Language Navigation (VLN), the formulation of prompts plays a critical role in determining evaluation metrics such as the Success Rate (SR). To ensure transparency and fairness in our experiments, we publicly release all prompts used in Table~\ref{tab:prompts}. It should be noted that, in addition to prompt formulation, factors including the frame rate of the simulator platform and the latency of online model invocation may also influence SR. The central aim of this work, however, is to introduce a highly challenging and promising new task based on the Airsim360 platform. Therefore, our experimental design prioritizes the most impactful factor, namely the formulation of prompts, while a comprehensive analysis of other variables remains outside the scope of this study.

\end{document}